\documentclass[preprint,12pt,nonatbib]{elsarticle}
\usepackage{graphicx} 

\usepackage[bb=dsserif]{mathalpha}

\usepackage{hyperref}
\usepackage{enumitem}
\usepackage{lscape}
\usepackage{amsfonts,amsmath,amsthm}
\usepackage{cleveref}
\usepackage{float}
\usepackage{booktabs}
\usepackage{bigints}
\usepackage{multirow}
\usepackage{makecell}
\usepackage{amssymb}
\usepackage{pifont}
\usepackage{xcolor}
\usepackage{graphicx}
\usepackage{epstopdf}
\usepackage{overpic}

\usepackage{mathtools}

\DeclarePairedDelimiter{\norm}{\lVert}{\rVert}

\newcommand*{\pijoint}{\boldsymbol{\pi}}

\newcommand*{\xjoint}{\mathbf{x}}
\newcommand*{\ujoint}{\mathbf{u}}

\newcommand*{\shield}{\boldsymbol{\Pi}}
\newcommand*{\pibackup}{\pijoint{}_{\textrm{backup}}}

\newcommand*{\XSet}{\mathcal{X}}
\newcommand*{\XSetSafe}{\mathcal{S}}
\newcommand*{\XSetCI}{\mathcal{X}_{\mathrm{CI}}}
\newcommand*{\XSetFI}{\mathcal{X}_{\mathrm{FI}}}

\makeatletter
\let\c@author\relax
\makeatother

\usepackage[
backend=biber,
style=numeric,
sorting=none
]{biblatex}
\bibliography{refs}

\newtheorem{Definition}{Definition}
\newtheorem{Theorem}{Theorem}
\newtheorem{Remark}{Remark}
\newcommand*{\argmin}{\textrm{argmin}}

\newcommand*{\cmark}{\ding{51}}\newcommand*{\xmark}{\ding{55}}

\definecolor{cmarkblue}{HTML}{1E88E5}
\definecolor{cmarkred}{HTML}{E53935}

\newcommand*{\cmarkB}{\textcolor{cmarkblue}{\cmark}}\newcommand*{\cmarkR}{\textcolor{cmarkred}{\cmark}}\newcommand*{\xmarkB}{\textcolor{cmarkblue}{\xmark}}\newcommand*{\xmarkR}{\textcolor{cmarkred}{\xmark}}

\definecolor{figorange}{HTML}{f4b183}
\definecolor{figgray}{HTML}{c9c9c9}
\definecolor{figblue}{HTML}{8eb4d1}

\newcolumntype{P}[1]{>{\centering\arraybackslash}p{#1}}

\journal{Annual Reviews in Control}

\begin{document}

\begin{frontmatter}

\title{Learning Safe Control for Multi-Robot Systems: Methods, Verification, and Open Challenges}

\author{Kunal Garg\corref{cor1}}
\cortext[cor1]{Corresponding author}
\ead{kgarg@mit.edu}

\author{Songyuan Zhang}\author{Oswin So}\author{Charles Dawson}\author{Chuchu Fan}

\affiliation{organization={Department of Aeronautics and Astronautics, Massachusetts Institute of Technology},addressline={77 Massachusetts Avenue}, 
            city={Cambridge},
            postcode={02139}, 
            state={MA},
            country={USA}}

\begin{abstract}
In this survey, we review the recent advances in control design methods for robotic multi-agent systems (MAS), focussing on learning-based methods with safety considerations. 
We start by reviewing various notions of safety and liveness properties, and modeling frameworks used for problem formulation of MAS. Then we provide a comprehensive review of learning-based methods for safe control design for multi-robot systems. We start with various types of shielding-based methods, such as safety certificates, predictive filters, and reachability tools. Then, we review the current state of control barrier certificate learning in both a centralized and distributed manner, followed by a comprehensive review of multi-agent reinforcement learning with a particular focus on safety. Next, we discuss the state-of-the-art verification tools for the correctness of learning-based methods. 
Based on the capabilities and the limitations of the state of the art methods in learning and verification for MAS, we identify various broad themes for open challenges: how to design methods that can achieve good performance along with safety guarantees; how to decompose single-agent based centralized methods for MAS; how to account for communication-related practical issues; and how to assess transfer of theoretical guarantees to practice. 
\end{abstract}

\begin{keyword}
Safe Multi-agent Reinforcement Learning; Certificate-based Multi-Agent Control; Verification for Multi-Agent Systems
\end{keyword}

\end{frontmatter}

\tableofcontents

\newpage

\section{Introduction}
\subsection{Motivation and applications of MAS}
Multi-agent systems (MAS) have received tremendous attention from scholars in different disciplines, including computer science and robotics, as a means to solve complex problems by subdividing them into smaller tasks \cite{dorri2018multi}. Some examples of MAS include smart grids \cite{ringler2016agent}, search and rescue teams \cite{vorotnikov2018multi,queralta2020collaborative}, edge computing \cite{wang2020multi}, wireless communication networks \cite{cui2019multi}, space systems \cite{ren2007distributed, wei2018learning, huang2023integrated}, package delivery \cite{salzman2020research}, power systems \cite{molzahn2017survey}, and micro-grids \cite{espina2020distributed}. The design and analysis of MAS controllers present unique challenges, such as scalability, verification, and robustness to factors such as communication issues, adversarial or non-cooperative agents, and partial observability. While we will provide a detailed discussion on the limitations and challenges of learning-based methods for MAS, interested readers on current broad challenges in various aspects of MAS are referred to \cite{ismail2018survey,canese2021multi,du2021survey,nweye2022real}. 

\subsection{Scope of this survey} This survey provides a comprehensive summary of the current state-of-the-art learning-based methods for safe MAS control. The four major topics of focus in this survey are
\begin{description}
    \item[Shielding-based methods --- Section \ref{sec: shielding}:] Methods that delegate safety to shielding function or safety filter which preserves the safety of the MAS. These include non-learning based Control Barrier Function (CBF), Predictive Safety Filters (PSF), Hamilton-Jacobi (HJ) Reachability, Automata-based methods and some heuristic methods.
    \item[Methods for learning CBF --- Section \ref{sec: cert learning}:] Methods for learning centralized and distributed CBF along with a control policy for MAS.
\item[Multi-agent reinforcement learning (MARL) --- Section \ref{sec: MARL}:] Methods that apply reinforcement learning to MAS and directly tackle safety constraints.
\item[Verification of learning-enabled MAS --- Section \ref{sec: MAS verification}:] The challenges inherent in verifying learned controllers for MAS, how various centralized verification tools have been extended to handle MAS, and communication issues specific to MAS.
\end{description}

\begin{figure}[h]
    \centering
    \includegraphics[width=1\columnwidth]{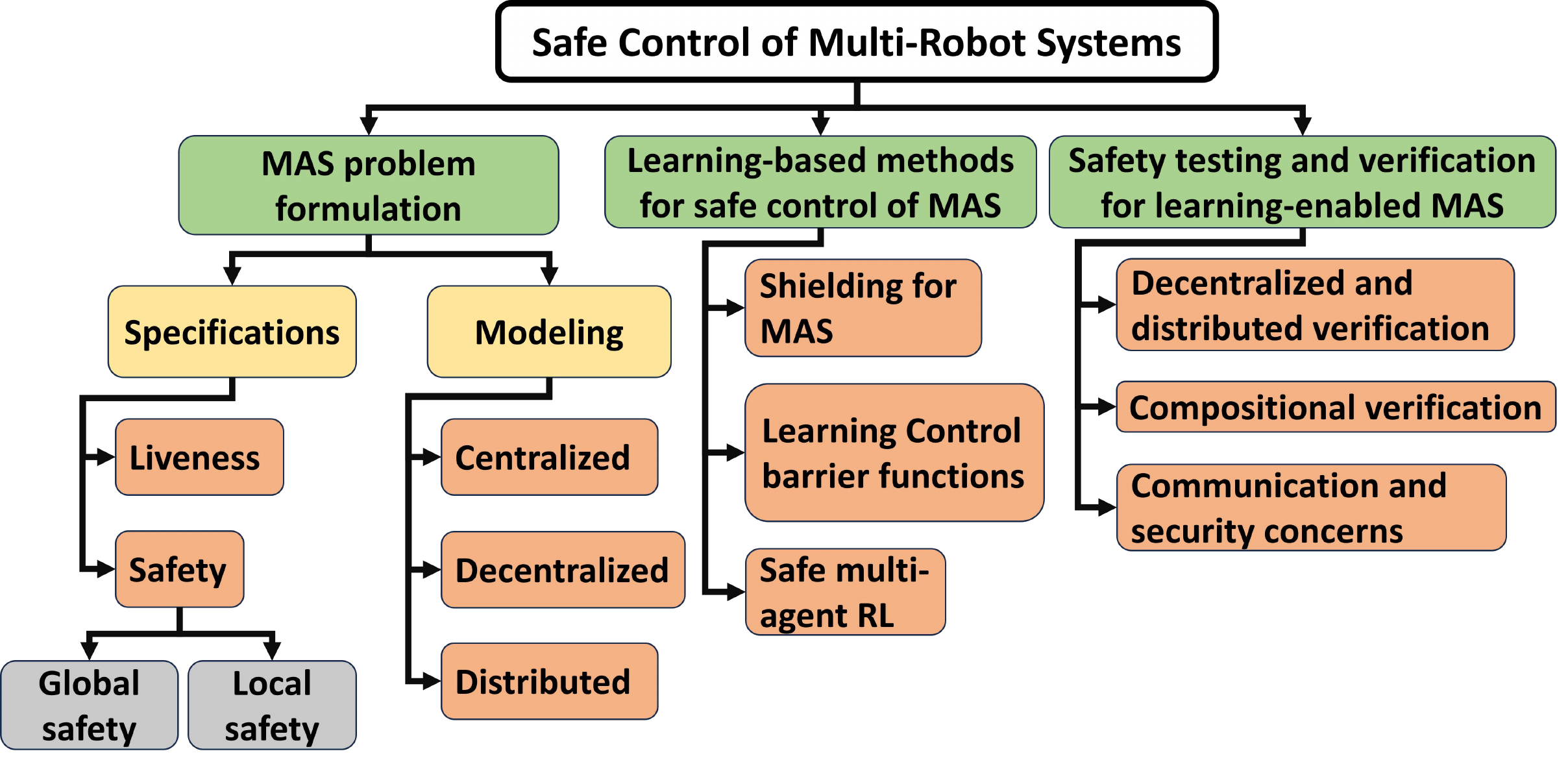}
    \caption{Overview of the survey taxonomy on safe learning for MAS.}
    \label{fig:overview}
\end{figure}

\subsection{Main takeaway from this survey}
\begin{itemize}
    \item \textit{Focus on MAS safety}: This is the first survey on robot MAS with an explicit focus on safety. The survey walks the reader through the taxonomy of the MAS control design problems, various learning-based methods for safe control synthesis, and the open problems in the field of safe MAS control. 
    \item \textit{Comprehensive survey of learning-based methods for multi-robot systems}: Unlike numerous existing surveys that only focus on MARL, this survey is intended to be a starting point for researchers getting started with learning-based safe control of MAS with a discussion of the capabilities and limitations of the existing learning-based tools.
    \item \textit{Vision for the future of safe MAS control}: The survey identifies a range of open problems in the field of safe learning-based control, and verification thereof, for robot MAS and it informs the future research on the topic. 
\end{itemize}

\subsection{Previous surveys on multi-agent systems}
Many survey and review articles appeared in the past few years on the topic of MAS. However, the key elements that differentiate this survey from the prior work are: 1) its focus on the safety of robotic MAS, 2) general learning methods as the central theme, and 3) its identification of open problems and challenges in safe learning-based MAS research. The article \cite{chen2019control} provides a detailed introduction to MAS taxonomy and control and \cite{tahir2019swarms} on Unmanned Aerial Vehicle swarms, but the discussion in these articles is restricted to non-learning-based methods. Given the popularity of MAS and safety in the context of RL, there have been many surveys that cover RL for MAS \cite{nguyen2020deep,zhang2021multi,zhang2021decentralized,oroojlooy2023review,zhou2023multi,rizk2019cooperative,gronauer2022multi} or for single-agent safety \cite{liu2021policy}.
Of these, only a few surveys \cite{zhou2023multi,gu2022review} cover the intersection of both topics. Despite this, many MARL works acknowledge safe MARL as a new area that has not been explored much but is a promising future direction \cite{zhang2021multi,oroojlooy2023review,gu2022review}. 

The topic of safety has been reviewed extensively in the robotics community \cite{brunke2022safe}. However, these works focus on the single-agent case, ignoring a broad category of disturbances and safety issues that are particular to MAS (e.g., communication delay and errors). While \cite{liu2021policy} discuss safety but for single-agent systems. There are quite a few surveys looking at distributed optimization \cite{molzahn2017survey,espina2020distributed,yang2019survey}, power systems \cite{molzahn2017survey} and micro-grids \cite{espina2020distributed}. While safety is not explicitly discussed in these works, some of the methods reviewed in these surveys can incorporate it via the inclusion of safety constraints. Finally, there are many surveys
focusing on applications of MAS \cite{queralta2020collaborative,xie2017multi,yang2019survey,espina2020distributed}. 

\subsection{Topics not covered by this article}
Given the main focus of the survey being safe learning-based methods for MAS, various topics pertaining to robotic MAS are out of the scope of this paper. A non-comprehensive list of such topics along with recent surveys on those topics is Vector field-based methods \cite{salman2017multi,gao2023non}; Model predictive control \cite{wang2020distributed}; Consensus control \cite{nowzari2019event}; Distributed optimization; \cite{yang2019survey,nedic2018distributed}; and Multi-agent games \cite{wang2022cooperative}. 

\subsection{Organization}
We start with a general problem formulation, notations, and common definitions for MAS in Section \ref{sec: MAS problem formulation}. Section~\ref{sec: shielding} introduces shielding methods for the safety of MAS, then Section \ref{sec: cert learning} discusses learned certificates more specifically.
Section \ref{sec: MARL} discusses various MARL-based methods. Section \ref{sec: MAS verification} covers verification techniques for MAS. 
Sections~\ref{sec: open probs} and~\ref{sec: conclusions} conclude with a discussion of the challenges and open problems in the field of safe MAS control. 

\section{MAS problem formulation}\label{sec: MAS problem formulation}
This section defines common notions used in the context of multi-agent systems (MAS); namely, agent model, specifications, and modeling frameworks. 

\subsection{Definitions and notations}\label{sec: def-and-notion}
In this work, we focus on a general class of MAS consisting of $N$ agents where each agent is a dynamic system modeled as 
\begin{align}\label{eq: MAS agent}
    x_i^+ = F(x_i, u_i, d_i, \nu_{ij}(x_j)), \quad x_i \in \mathcal X_i, \quad u_i\in \mathcal U_i,
\end{align}
where $x_i\in \mathbb R^{n_i}, u_i\in \mathbb R^{m_i}$ denote the state and the input of agent $i$ and $x^+ = (x[k+1] - x[k])$ if the agents are modeled as discrete-time systems with $k\in \mathbb N$ denoting the discrete time, and $x^+ = \dot x(t)$, for continuous-time systems with $t\in \mathbb R_+$ being the continuous time. The term $d_i\in \mathcal D_i$ denotes the disturbances, uncertainties, and unmodeled dynamics for agent $i$, while the map $\nu_{ij}:\mathbb R^{n_j} \to \mathbb R^{p_i}$ denotes the influence of the other agents $j\neq i$ or in other words, the inter-agent coupling of the MAS dynamics. The joint state vector and the input vector are denoted as $\mathbf x = [x_1^\top, x_2^\top, \dots, x_N^\top]^\top\in\mathbb R^{\sum n_i}$ and $\mathbf u = [u_1^\top, u_2^\top, \dots, u_N^\top]^\top\in\mathbb R^{\sum m_i}$, respectively. The combined dynamics of the MAS can be written as
\begin{align}\label{eq: MAS dyn}
    \dot {\mathbf x} = \mathbf F(\mathbf x, \mathbf u, \mathbf d).
\end{align}
To enable theoretical analysis, it is typically assumed that the function $\mathbf F$ is locally Lipschitz continuous \cite{ames2016control}. 

For robotic systems, let $x_i \supset p_i\in \mathbb R^3$ denote the physical location of the $i$-th agent in the 3D space. The sets $\mathcal X_i \subseteq \mathbb R^{n_i}, \mathcal U_i\subseteq\mathbb R^{m_i}$ denote the operational workspace and the set of inputs for the $i$-th agent.  The state trajectory of agent $i$ under a control policy $\pi_i$ starting at an initial condition $x_i(0)\in \mathcal X_i$ is denoted as $\phi_i(\cdot,\pi_i; x_i(0)):\mathbb R_+\to\mathbb R^{n_i}$. Correspondingly, the state trajectory of the MAS is denoted as $\Phi(\cdot, \pijoint; \mathbf x(0))$ with $\pijoint$ being the joint policy for the MAS. When an explicit emphasis on the underlying policy is not required, we denote the trajectories of the agents with $x_i(\cdot)$ and that of the MAS with $\mathbf x(\cdot)$ for the sake of brevity. 

A network can be defined for the MAS with agents denoting the nodes and their communication links denoting edges. Let $R_i>0$ be the sensing/communication radius of the $i$-th agent and define $\mathcal N_i(t) = \{j \; |\; \|p_i-p_j(t)\|\leq R\}$ as the set of \textit{neighbors} of the $i$-th agent, i.e., the set of agents from (respectively, to) which the agent $i$ can receive (respectively, send) information \cite{zavlanos2008distributed}. Using this notion of neighbors, a graph topology can be defined for the MAS as $\mathcal G(t) = (\mathcal V, \mathcal E(t))$ where $\mathcal V = \{1, 2, \dots, N\}$ is the set of vertices denoting the agents and $\mathcal E(t)$ the time-varying set of connections given as $\mathcal E(t) = \{(i, j)\; :\; j\in \mathcal N_i, i\in \mathcal V\}$, that is, there is an edge $\mathcal E_{ij}(t)$ from agent $j$ to agent $i$ at time $t$ if the agent $i$ is able to receive information from agent $j$. A time-varying adjacency matrix $\mathcal A$ for this graph is defined as
\begin{align}
    \mathcal A_{ij}(t) = \begin{cases}
        1 & j\in \mathcal N_i(t), \\
        0 & \text{otherwise}.
    \end{cases}
\end{align}
The Laplacian matrix $\mathcal L$ corresponding to the adjacency matrix $\mathcal A(t)$ is given as
\begin{align}
    \mathcal L_{ij}(t) = \begin{cases}
        \sum\limits_{k\neq i}\mathcal A_{ik}(t), & j = i, \\
        0 , & \text{otherwise}.
    \end{cases}
\end{align}
From \cite[Theorem 2.8]{mesbahi2010graph}, we know that the graph topology $\mathcal G(t)$ is connected at time $t$ if and only if the second smallest eigenvalue of the Laplacian matrix is positive, i.e., $\lambda_2(\mathcal L(\mathcal A(t))) > 0$. 

\subsection{MAS specifications}
In the temporal logic language \cite[Chapter 3]{baier2008principles}, the control objective for the MAS can be characterized as:
\begin{enumerate}
    \item \textbf{Safety property}: \textit{Something bad never happens}. Agents remain in the safe region $\mathcal S(t)\subset \mathbb R^{\sum n_i}$ at all times, i.e., $\mathbf x(t)\in \mathcal S(t)$ for all $t\geq 0$ \cite{wang2016safety,zhang2019mamps}. Generally, the safe region is defined as the complement of the occupancy set of other agents, obstacles, and restricted regions in the workspace.
    \item \textbf{Liveness property}: \textit{Something good will eventually happen}. Agents move towards minimizing\footnote{In some works, the objective is given as maximization instead of minimization.} a (possibly joint) objective function $\Psi:\mathbb R^{\sum n_i} \to \mathbb R$, i.e., $\mathbf x(t) \to \argmin_{\mathbf {x}} \Psi(\mathbf x)$ \cite{atincc2020swarm,prajapat2022near,zhang2018fully,sun2022multi,chen2021scalable}. The most common example of a liveness property is each agent required to reach a goal location. 
\end{enumerate}
Some safety properties can be decomposed at the agent level, while some safety properties need to be stated for the MAS as a whole. For example, the inter-agent safety property needs to be specified for the MAS using a \textit{global} safe set $\mathcal S$ while obstacle avoidance property can be expressed individually for each agent with a \textit{local} safety set $\mathcal S_i$. Another important requirement that can be posed as a safety property in MAS is \textit{connectivity} maintenance, which becomes an important property for various applications such as coverage \cite{cortes2004coverage} and formation control \cite{mehdifar2020prescribed}. For the sake of performance criteria as well as maintaining safety, the agents in MAS need to sense each other and might also need to actively communicate certain information. A detailed discussion on various safety properties is provided next. 

Note that there are various notions of safety used in the literature, however, since the focus of this survey is robotic MAS, we restrict our discussions to safety as it pertains to \textit{physical} safety of the agents.
\subsubsection{Safety property}
For a MAS \eqref{eq: MAS agent}, the notion of safety is defined as follows. 
\begin{Definition}[\textbf{Safety}] \label{def:mas_safety} Given a (potentially time-varying) safety constraint set $\mathcal S(t)$, the MAS \eqref{eq: MAS agent} is safe with respect to $\mathcal S(t)$ if for all $\mathbf x(0)\in \mathcal S(0)$, the trajectories satisfy $\mathbf x(t)\in \mathcal S(t)$ for all $t\geq 0$. 
\end{Definition}

Below, we give examples of some of the possible safety constraints for robotic MAS. We draw a distinction between properties that require considering the joint state of the MAS (global properties) and those that can be checked using only individual agent states (local properties).
\begin{enumerate}
\item Global MAS safety properties
\begin{itemize}
    \item \textit{Inter-agent collision avoidance}: Given a safe distance $0<r_s<R$, the inter-agent collision avoidance can be formulated through $\mathcal S = \{\mathbf x\; |\; \|p_i - p_j\|> r_s, j \neq i\}$ \cite{panagou2016distributed,zhang2023neural}. 
    \item \textit{Connectivity maintenance}: Given a communication radius $R>0$, the MAS network connectivity maintenance can be formulated as $\mathcal S = \{\mathbf x \; |\; \lambda_2(\mathcal L(\mathcal A(\mathbf x))) > 0\}$ \cite{zavlanos2008distributed,sabattini2013distributed}, where 
    \begin{align}
        \mathcal A_{ij}(\mathbf x) = \begin{cases} 1 , & \|p_i - p_j \|\leq R, \\
        0, & \text{otherwise}.
        \end{cases}
    \end{align} 
\end{itemize}
\item Local safety properties
\begin{itemize}
    \item \textit{Obstacle avoidance}: Given a safe distance $0<r_s<R$ and a set of (potentially moving) obstacles $\mathcal O_j(t)\subset \mathbb R^3$, obstacle avoidance can be formulated through the set $\mathcal S_i(t) = \{p_i \; |\; \argmin_{p\in \mathcal O_j(t)}\|p_i-p\| > r_s, \forall j\}$ \cite{zhang2023neural,chen2020guaranteed}.
\item \textit{State limits}: Given state limits (e.g., velocity, acceleration, torque limits) of the form $x_m^j\leq x_i^j\leq x_M^j$ where $x_i^j$ denotes the $j-$th component of $x_i$, the safety can be formulated with the set $\mathcal S_i = \{x_i \; |\; x^j_m\leq x_i^j\leq x^j_M, j \in \{1, 2, \dots, n_i\}\}$\footnote{More general constraints of the form $r(x_i) \leq 0$ for some constraint function $r$ can also be considered.} \cite{borrmann2015control,xian2017closed}.
\end{itemize}
\end{enumerate}

\subsubsection{Liveness properties} \label{subsubsec:liveness}
Here, we discuss commonly studied liveness properties for a MAS. \footnote{Some of these properties require compatible workspaces for the agents, i.e., $n_i = n_j = n$ for all $i$.}:
\begin{enumerate}
    \item \textit{Consensus}: Given a consensus point $\rho(x_c)$ for some function $\rho$, the trajectories of each of the agents converge in the following sense: $\lim\limits_{t\to\infty}\rho(x_i(t)) = \rho(x_c)$ \cite{li2020consensus,ren2005survey}. 
    \item \textit{Formation}: Given a set of off-set vectors $x_{ij}$ for each $(i,j), \; i\neq j$, the trajectories of the MAS satisfy $x_i(t) - x_j(t) \to x_{ij}$ as $t\to \infty$ \cite{oh2015survey,xue2019leader}.
    \item \textit{Coverage}: Given an exploration region $\mathcal X_e\subset \mathcal X$ and a distribution function $\varphi:\mathcal X_e \to \mathbb R_+$ and a smooth increasing function $f:\mathbb R\to \mathbb R$ that measures the degradation of sensing performance, the coverage objective is to minimize the function $\Psi(x) = \sum\limits_{i = 1}^N \bigintssss_{V_i}f(\|x_i-z\|)\phi(z)dz$, where $V_i\subset \mathcal X_e$ denotes the region where agent $i$ is responsible for coverage \cite{cortes2004coverage,santos2018coverage}.
    \item \textit{Goal reaching}: Given a set of goal states $x_{gi}\in \mathbb R^{n_i}$, the trajectories of each agent reach the goal, i.e., $\lim\limits_{t\to \infty}x_i(t) = x_{gi}$ for each $i$ \cite{majumdar2021symbolic,panagou2016distributed,garg2019finite}.
    \item \textit{Reference tracking}: Given a reference trajectory $x_{i, ref}(\cdot)$, the trajectories of each of the agent satisfy $x_i(t) - x_{i, ref}(t) \to 0$ as $t \to \infty$ \cite{adaldo2016multi,saim2017distributed}.  
\end{enumerate}

The liveness properties are important for capturing both performance criteria (e.g. goal reaching or trajectory tracking) and certain safety properties (particularly global safety properties). For example, MAS might be required to reach a consensus or maintain a formation in order to maintain a connectivity property. While the focus of this survey is on safety properties, we provide brief comments on the ability of the reviewed methods to accomplish tasks that often require both safety and liveness properties (particularly, goal-reaching). 

\subsection{MAS modeling framework}

\begin{figure}
    \centering
    \includegraphics[width=1\columnwidth]{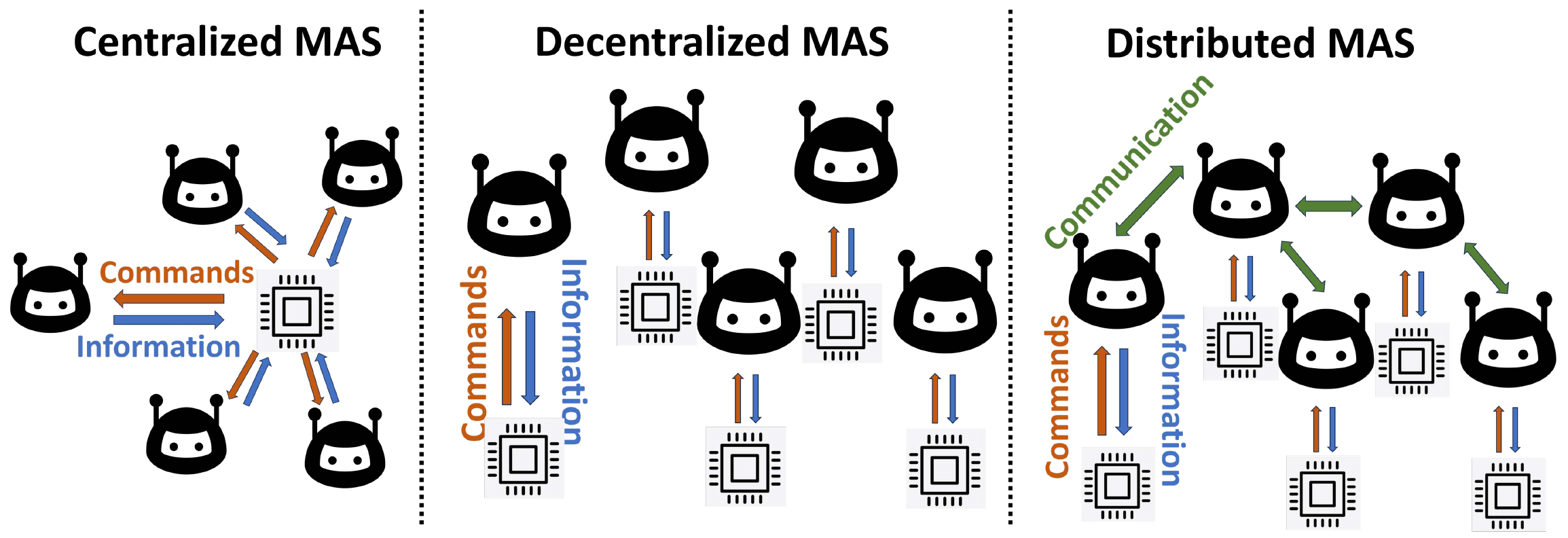}
    \caption{MAS modeling: centralized, decentralized and distributed frameworks. }
    \label{fig:central-decentral-distribted}
\end{figure}

\subsubsection{Centralized, decentralized, and distributed MAS}
In this section, we present various modeling frameworks for MAS. Generally, MAS is modeled under the following three paradigms:
\begin{enumerate}
    \item \textit{Centralized}: An MAS is termed centralized if there is a \textit{central} node where all the information/sensor data from all the agents is collected and decisions for each of the agent is made. 
    \item \textit{Decentralized}: An MAS is termed decentralized if each agent makes its own decision based on its local information/sensor data \textit{without} communicating with other agents.
    \item \textit{Distributed}: An MAS is termed distributed if each agent makes its own decisions based on its local information/sensor data along with information received by active communication with other agents. 
\end{enumerate}

We note there is currently no consensus in the literature on the definition of the decentralized and distributed paradigms. For example, \cite{xuan2002multi} defines decentralized MAS where the agents communicate with their local neighbors. The authors in \cite{roth2005decentralized} use a framework where agents communicate only when \textit{needed}, calling this approach decentralized communication. In more recent multi-agent reinforcement learning (MARL) work \cite{zhang2018fully}, the authors allow the agents to communicate over a time-varying connectivity graph and call the formulation \textit{fully decentralized}. There are many examples of this interchangeable usage of the terms decentralized and distributed in the literature. For the sake of consistency, in this survey, we will stick with the notions as defined above (see \cite{frampton2010comparison,molzahn2017survey}). 

Some examples of methods using centralized learning frameworks for safe control of MAS are \cite{zhang2019mamps,elsayed2021safe,khan2019learning,gu2021multi,dawson2022clbf}, the works in \cite{cai2021safe,melcer2022shield} use a decentralized learning framework, and \cite{lu2021decentralized,pereira2022decentralized} use a distributed learning framework. Centralized Training Decentralized Execution (CTDE) is a related paradigm, where the joint state and other global information are used to train a decentralized policy for each agent that only has access to local information \cite{zhang2023neural,gronauer2022multi}.

\begin{table}[h]
\centering
\begin{tabular}{@{}p{.14\linewidth}P{.11\linewidth}P{.11\linewidth}P{.14\linewidth}P{.14\linewidth}P{.19\linewidth}@{}}\toprule
            & \multicolumn{2}{c}{Safety Guarantees} & \multicolumn{2}{c}{Requirements} & \\ \cmidrule(lr){2-3}\cmidrule(lr){4-5}
    Method  & \makecell{Theory} & \makecell{Practice} & \makecell{Domain\\ Expertise} & \makecell{Known\\ Dynamics} & \makecell{Distributed\\Policy} \\ \midrule
Shielding & \cmarkB & \cmarkB & \xmarkB/\cmarkR\footnotemark & \cmarkR & \cmarkB/\xmarkR\footnotemark \\\addlinespace[0.9em]
    \makecell[l]{Certificate\\ Learning} & \cmarkB & \xmarkR & \xmarkB & \cmarkR & \cmarkB \\\addlinespace[0.9em]
    \makecell[l]{Unconstrained\\MARL} & \xmarkR & \xmarkR & \xmarkB & \xmarkB & \cmarkB \\\addlinespace[0.5em]
    \makecell[l]{Constrained\\MARL}   & \cmarkB & \xmarkR & \xmarkB & \xmarkB & \cmarkB \\
\bottomrule
\end{tabular}
\caption{Overview of different methods of handling safety for MAS. The tick mark denotes available features or requirements, while the cross mark denotes missing features or non-requirements. Blue denotes desirable properties, while red denotes undesirable properties.
}
\label{tab:methods_overview}
\end{table}

\subsection{Properties of algorithms for MAS Safety}
Finally, we discuss some desirable properties of learning-based algorithms for the safe control of MAS and categorize the main themes reviewed in this paper based on these properties (see Table \ref{tab:methods_overview}):
\begin{description}
    \item[Safety Guarantees - Theory: ] Here, we categorize the methods based on the fact that whether, under some suitable assumptions, it results in a safe policy.
    \item[Safety Guarantees - Practice: ] While theoretical guarantees are important, it is more useful to ask whether the assumptions needed to provide those safety guarantees from theory also hold in practice.    
\end{description}

\footnotetext[5]{HJ and Automata-based methods can be applied directly from safety specifications, while PSF (needs valid control-invariant set) and CBF-based (needs a valid CBF) shielding require domain expertise to use.}
\footnotetext[6]{CBF \cite{cai2021safe}, PSF \cite{muntwiler2020distributedmpc} and automata-based \cite{melcer2022shield} shielding have distributed versions that maintain their safety guarantees in practice. While HJ-based shielding does have a distributed version by considering pairwise interactions \cite{chen2015hj}, the safety guarantees do not hold for the full MAS.}

\begin{description}
    \item[Requirements - Domain Expertise: ] An important aspect that dictates the ease of usage and wide applicability of a method is whether domain expertise is needed about the specific dynamics or safety constraints to construct supporting tools and methods needed to apply the method.
    \item[Requirement - Known Dynamics: ] Another factor that plays an important role in the generalizability and wide applicability of a method is whether the exact form of the dynamics is needed (e.g., for computing jacobians / querying at arbitrary states) to apply the method, or it is sufficient to have black-box evaluations of the dynamics along a set of trajectories.
    \item[Distributed Policy: ] Finally, and very importantly for large-scale MAS, one needs to ask whether the policy can be deployed in a distributed manner without a central computation/aggregation node.
\end{description}

\section{Shielding-based learning for MAS}\label{sec: shielding}
One popular method of providing safety to learning-based methods is via the use of \textit{shielding} or  \textit{safety filter}s, where an \textit{unconstrained} learning method is paired with a \textit{shield} or \textit{safety filter}. Such shields are often constructed without learning with the objective of either modifying the input or the output of the learning method to maintain safety. One benefit of shielding-based methods is that safety can be guaranteed during both training and deployment since the shield is constructed prior to training. However, a drawback of some of these methods is that they require domain expertise for the construction of a valid shield, which can be challenging in the single-agent setting and becomes even more difficult for MAS. Other methods can automatically synthesize shields, but face scalability challenges.

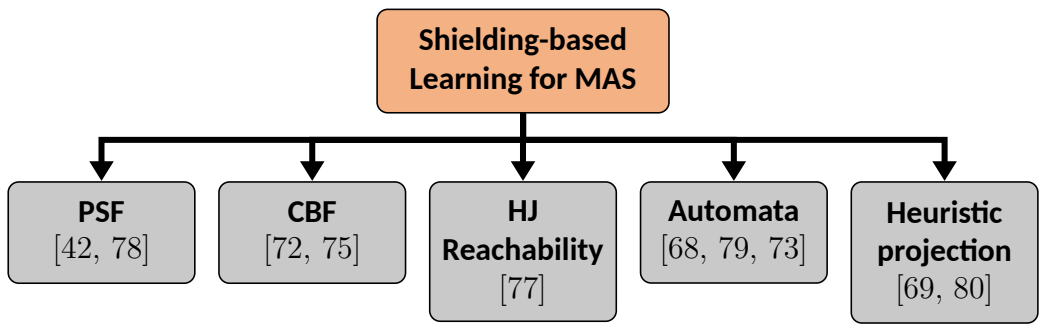
\begin{figure}[h]
\centering
\begin{overpic}[width=\linewidth,abs,unit=1mm,scale=.25]{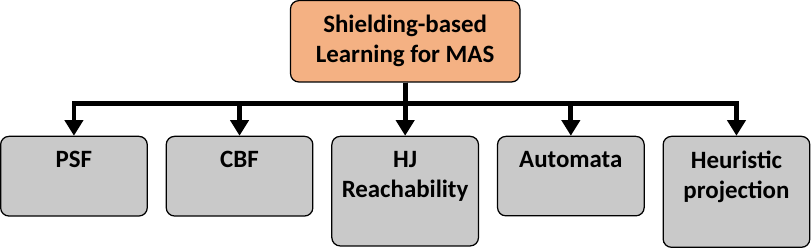}
  \put(6,8.715){\cite{zhang2019mamps,vinod2022safe}}
  \put(34,8.715){\cite{cai2021safe,pereira2022decentralized}}
  \put(65.3,3.45){\cite{chen2015hj}}
  \put(86.7,8.715){\cite{elsayed2021safe,xiao2023model,melcer2022shield}}
  \put(118,3.45){\cite{khan2019learning,sheebaelhamd2021safe}}
\end{overpic}
\caption{Overview of Shielding-based Learning for MAS.}
\label{fig:shielding_overview}
\end{figure}

Let $\pijoint: \mathcal{X} \to \mathcal{U}$ denote the \textit{joint} policy for MAS \eqref{eq: MAS dyn} from a learning-based controller without safety considerations. Shielding-based methods define a \textit{shielding function} $\shield: \mathcal{X} \times \mathcal{U} \to \mathcal{U}$ that takes the output of $\pijoint$ and returns a shielded output \cite{zhang2019mamps}. The level of safety and the type of safety guarantees that can be obtained depends on how the shielding function $\shield$ is constructed. We provide an overview of different shielding-based methods used to ensure the safety of learning for MAS in Figure \ref{fig:shielding_overview}.

 \subsection{Control barrier function-based shielding}\label{sec: cbf-shield}
One method of constructing a shield $\shield$ is via a Control Barrier Function (CBF). We start by reviewing the notion of CBF. 

\subsubsection{Definition of CBF}

The notion of CBF was introduced to satisfy the conditions of set invariance, where a set is termed as \textit{forward invariant} if starting in the set, the system trajectories do not leave it \cite{nagumo1942lage,blanchini1999set,brezis1970characterization}. It is also related to the notion of Control Lyapunov Function (CLF) \cite{sontag1983lyapunov,ames2014rapidly} which is commonly used for liveness properties, and extends the definition of barrier certificates \cite{prajna2004safety,prajna2006barrier} to control systems. A comprehensive review of CBFs as a tool for safety can be found in \cite{ames2019control}. 

While there exists various definitions of CBF with slight variations\cite{wieland2007constructive,ames2014control,ames2016control,dawson2022safe}, we use the following definition in this survey \cite{ames2019control}: 

\begin{Definition}[\textbf{CBF}]
Consider the MAS dynamics $\eqref{eq: MAS dyn}$ with no disturbances, i.e., $\mathbf d=0$. 
Let $\mathcal C\subset\mathcal X$ be the $0$-superlevel set of a continuously differentiable function $B: \mathcal X\rightarrow\mathbb R$, i.e., $\mathcal C= \{\mathbf x\in \mathcal X: B(x) \geq 0\}$. Then, the function $B$ is a CBF if there exists an extended class-$\mathcal{K}_\infty$ function $\alpha: \mathbb R\rightarrow\mathbb R$ \footnote{A continuous function $\alpha: \mathbb R\rightarrow\mathbb R$ is said to be an extended class-$\mathcal{K}_\infty$ function if it is strictly increasing with $\alpha(0)=0$ and $\lim\limits_{s\to \infty}\alpha(s) = \infty$.} such that:
\begin{equation}\label{eq: cbf-descent-cond}
    \sup_{\mathbf u\in\mathcal U} \left[\frac{\partial B}{\partial \mathbf x}\mathbf F(\mathbf x,\mathbf u) + \alpha\left(B(\mathbf x)\right)\right]\geq 0, \quad \forall \mathbf{x}\in\mathcal X. 
\end{equation}
\end{Definition}
\textit{Safe Control Set:} Based on \eqref{eq: cbf-descent-cond}, we can define a set of \textit{safe} control input 
\begin{equation}
    K_\mathrm{CBF}(\mathbf x) = \left\{\mathbf u\in\mathcal U: \frac{\partial B}{\partial \mathbf x}\mathbf{F}(\mathbf x,\mathbf u) + \alpha\left(B(\mathbf x)\right)\geq 0\right\}.
\end{equation}
The authors in \cite{ames2019control} proved the following result on forward invariance of the set $\mathcal C$ using the notion of CBF:

\begin{Theorem}\label{thm: cbf-safety}
    Let $\mathcal{C}$ be the $0$-superlevel set of a continous differentiable function $B: \mathcal X\rightarrow\mathbb R$, i.e., $\mathcal C = \{\mathbf x\in\mathcal X: B(\mathbf{x})\geq 0\}$. 
    If $B$ is a CBF, and $\frac{\partial B}{\partial \mathbf x}\ne0$ for all $\mathbf x\in\partial \mathcal C$, then any Lipschitz continuous policy $\pijoint:\mathcal X \to \mathcal U$ with $\pijoint(\mathbf x)\in K_\mathrm{CBF}(\mathbf x)$ renders the set $\mathcal C$ forward invariant. 
\end{Theorem}

\begin{Remark}\label{rmk: cbf-safety}
    The forward invariance of the set $\mathcal C$ can be used for guaranteeing safety for a set $\mathcal S$ as follows. 
    Based on \Cref{thm: cbf-safety}, if a CBF and a controller are found on $\mathcal X$ that satisfy the conditions of \Cref{thm: cbf-safety} and $\mathcal C \subset \mathcal S$, then starting from any initial condition in $\mathcal C$, the system remains safe. 
\end{Remark}

There are variations of CBF that can also render the safety of autonomous systems. For example, an appropriate choice of Lyapunov function can be used for safety since the sublevel sets of a Lyapunov function are forward invariant \cite{tee2009barrier}. Based on this idea, \cite{dawson2022clbf} combined CLF and CBF, and introduced a framework for learning a Control Lyapunov Barrier Function (CLBF) that guarantees both safety and stability. The approach in \cite{yu2023learning} proposes another variation of CBF called the Control Admissibility Model (CAM), which uses a notion of CBF where the function $B$ depends on both the state $\mathbf x$ and the control input $\mathbf u$. The central idea of most of the variations is still forward invariance as per \Cref{thm: cbf-safety}.

\subsubsection{CBF-based shielding synthesis}

For control-affine systems $\dot{\mathbf x}=\mathbf{F}(\mathbf x)+\mathbf{G}(\mathbf x)\mathbf u$, condition \eqref{eq: cbf-descent-cond} becomes a linear constraint in the control input $\mathbf u$. When the input constraint set $\mathcal U$ is a convex polytope, a centralized method to synthesize a safe control input through CBF-based shielding is using Quadratic Programming (QP):
\begin{equation}\label{eq: cbf-qp}
    \begin{aligned}
        \min_{\mathbf u\in\mathcal U}\quad & \|\mathbf u-\boldsymbol{\pi}_\mathrm{nom}(\mathbf x)\|^2,\\
        \mathrm{subject\;to}\quad & L_\mathbf{F} B(\mathbf x) + L_\mathbf{G} B(\mathbf x)\mathbf u + \alpha\left(B(\mathbf x)\right)\geq 0,
    \end{aligned}
\end{equation}
where $L_\mathbf{F} B(\mathbf x)=\nabla B(\mathbf x)^\top\mathbf{F}(\mathbf x)$ and $L_\mathbf{G} B(\mathbf x)=\nabla B(\mathbf x)^\top\mathbf{G}(\mathbf x)$ and $\boldsymbol{\pi}_\mathrm{nom}: \mathcal X\rightarrow\mathcal U$ is a nominal policy (from some uncontrained learning methods) that does not necessarily consider safety with respect to the set $\mathcal C$.
Given a state $\xjoint$ and a control policy $\pijoint_{\mathrm{nom}}$, the CBF-based shield $\shield$ outputs the solution $\ujoint^*$ to the QP \eqref{eq: cbf-qp} that minimaly modifies the nominal control input while guaranteeing the safety of the system. Here, the nominal policy may come from a learning-baed approach without safety considerations, and the QP \eqref{eq: cbf-qp} provides a shielding mechanism of enforcing safety with this learned control policy. 

\subsubsection{Constructing CBF}

Once a CBF is given for control-affine systems, we can solve problem \eqref{eq: cbf-qp} for shielding. However, finding a CBF for MAS is not trivial, and there is no generalized framework that can find a CBF for any MAS. Here, we review approaches that can efficiently compute CBFs for certain specific types of systems. 

For systems with relatively simple dynamics, such as single integrator, double integrator, and unicycle dynamics, it is possible to use a distance-based CBF \cite{ames2014control,ames2016control,wu2016safety,xu2017realizing,ames2019control,garg2019control,garg2021robust,garg2022fixed,yin2023shielding,tong2023enforcing}. Some works also explore provable safety along with liveness by guaranteeing the feasibility of the underlying CBF-QP \cite{garg2021robust,garg2022fixed,garg2021multi}. For systems with multiple safety constraints, e.g., velocity constraints, and joint angle constraints, it is possible to design a CBF for each constraint and then combine them \cite{hsu2015control,glotfelter2017nonsmooth,usevitch2020strong}. However, one needs domain expertise to handcraft each of these CBFs. It is also difficult to encode input constraints when handcrafting the CBF, and therefore, the CBF-QP \eqref{eq: cbf-qp} can be infeasible. 

For systems with polynomial dynamics, it is possible to use the Sum-of-Squares (SoS) method to compute a CBF. The key idea of SoS is that the CBF conditions \eqref{eq: cbf-descent-cond} consists of a set of inequalities, which can be equivalently expressed as checking whether a polynomial function is SoS. In this manner, a CBF can be computed through convex optimization \cite{ahmadi2016some,clark2021verification,srinivasan2021extent}. However, these methods are limited to polynomial dynamics. Moreover, the SoS-based approaches suffer from the curse of dimensionality (i.e., the computational complexity grows exponentially with respect to the degree of polynomials involved) \cite{ahmadi2016some}. 

\subsubsection{Distributed CBF}

While centralized CBF is an effective shield for small-scale MAS, due to its poor scalability, it is difficult to use it for large-scale MAS. To address the scalability problem, the notion of distributed CBF can be used \cite{zhang2023neural,lindemann2019control,panagou2013multi,cai2021safe}. In contrast to centralized CBF where the state $\mathbf x$ of the MAS is used, for a distributed CBF, only the local observations and information available from communication with neighbors are used, reducing the problem dimension significantly. 

Similar to a variety of notions of centralized CBF, there exists a variety of definitions of distributed CBF in the literature \cite{borrmann2015control,wang2017safety,glotfelter2017nonsmooth,lindemann2019control,qin2020learning,zhang2023neural}. Following the graph notions of MAS introduced in \Cref{sec: def-and-notion}, we review a slightly modified definition of distributed CBF from \cite{zhang2023neural}. Let $o_i\in\mathcal O_i \subset \mathbb R^{r_i}$ be the observation vector of agent $i$, and let $z_i\in\mathcal Z_i \subset \mathbb R^{q_i}$ be the encoding\footnote{There are many ways of encoding information, such as concatenation \cite{borrmann2015control}, summation \cite{qin2020learning}, and applying attention \cite{zhang2023neural}.} of the information accepted by agent $i$ from its neighbors $\mathcal N_i$. Note that $z_i$ depends on the states of the neighbors of agent $i$.

\begin{Definition}[\textbf{Distributed CBF}]\label{def: distributed CBF}
    Consider the MAS agent dynamics \eqref{eq: MAS agent} with no disturbances and no inter-agent influences, i.e., $d_i=0$ and $\nu_{ij}(x_j)=0$, for all $i, j\in \mathcal V$. Let $\mathcal C_i\subset\mathcal X_i$ be the $0$-superlevel set of a continuously differentiable function $B_i:\mathcal X_i\times \mathcal O_i\times \mathcal Z_i\rightarrow\mathbb R$. Then, the function $B_i$ is a distributed CBF if there exists an extended class-$\mathcal K_\infty$ function $\alpha_i: \mathbb R\rightarrow\mathbb R$ and for each $\mathbf x \in \mathcal X$, there exists a control input $\mathbf u\in\mathcal U$, such that the following holds:
    \begin{align}\label{eq: distributed-cbf-descent}
\frac{\partial B_i}{\partial x_i}F_i(x_i,u_i)+\frac{\partial B_i}{\partial o_i}\dot o_i
            +\sum_{j\in\mathcal N_i\cup \{i\}}\frac{\partial B_i}{\partial z_i}\frac{\partial z_i}{\partial x_j}F_j(x_j,u_j)+\alpha_i(B_i(x_i))\geq 0,\quad\forall i.
\end{align}
\end{Definition}

Similar to the centralized CBF, under certain conditions, the distributed CBF can also guarantee the safety of the MAS \cite{zhang2023neural}. Accoding to Definition \ref{def: distributed CBF}, the MAS is assumed to be cooperative, i.e., all the agents coordinate to satisfy CBF condition \eqref{eq: distributed-cbf-descent} for MAS \cite{wang2016safety,wang2017safety,qin2020learning,machida2021consensus,zhang2023neural}. There are also other works that consider the worst-case scenario that the agents are non-cooperative \cite{borrmann2015control}, in which case, the agents do not have any communication, resulting in a decentralized CBF formulation. 

\subsubsection{Constructing distributed CBF}

One way to construct distributed CBF is by \textit{decomposing} centralized CBF. There are many ways to decompose centralized CBF. For example, assuming that other agents keep constant velocities, actively chasing the \textit{ego} agent, or actively avoiding collision with the ego agent \cite{borrmann2015control}. Other works \cite{wang2016safety,wang2017safety,lindemann2020barrier,cosner2023learning} consider risk allocation among agents while decomposing the centralized CBF. In addition, 
decomposing the centralized CBF allows each agent solve the optimization problem individually based on their local information in a distributed fashio, and therefore reduces computation costs \cite{borrmann2015control,wang2017safety,pereira2022decentralized}. 

Another way to construct a distributed CBF is through a bottom-up approach, e.g., by composing pair-wise CBF. These approaches often encode the constraints of all the pair-wise CBF conditions in a QP to find a feasible control input that can maintain safety with respect to each of the pair-wise CBF \cite{glotfelter2017nonsmooth,glotfelter2018boolean,funada2019visual,luo2020multi,chen2020guaranteed,jankovic2021collision,mali2021incorporating,hu2022decentralized,jiang2023incorporating,yu2023learning}. 

\subsection{Predictive safety filter-based shielding}
Predictive safety filter (PSF) is another common method of constructing a shield \cite{wabersich2018linear}, which is also closely related to model predictive shielding (MPS) \cite{li2020robust,bastani2021safe}. 

For PSF, let $\XSetCI \subseteq \XSetSafe$ denote a subset of the safe set that is \textit{control invariant}, i.e., there exists some controller under which this set is forward invariant. During deployment, at each timestep, a constrained optimization problem is solved that constrains the terminal state within $\XSetCI$.
\begin{subequations} \label{eq:psf:optim}
\begin{align}
    \min_{u_k} \quad &\norm{\pijoint(\xjoint_0) - \ujoint_0}^2 \\
    \textrm{s.t.} \quad
        & \xjoint_{k+1} = f(\xjoint_k, \ujoint_k), \quad \forall k = 0, 1, \dots, N - 1 \\
        & \xjoint_{k} \in \XSetSafe, \quad \forall k = 0, 1, \dots, N - 1 \\
        & \xjoint_{N} \in \XSetCI \label{eq:psf:term_constr}.
\end{align}
\end{subequations}
The output of the safety filter $\shield$ is then taken as the solution $\ujoint_0$ of \eqref{eq:psf:optim}.
The terminal state constraint \eqref{eq:psf:term_constr} helps guarantee recursive feasibility of \eqref{eq:psf:optim} and, as a result, guarantees infinite-horizon constraint satisfaction.

In MPS, suboptimality of the underlying constrained optimization problem is traded for substantially lowered computational costs.
Instead of solving the potentially nonlinear optimization problem \eqref{eq:psf:optim} at each timestep, a merely feasible solution is found. Let $\pibackup$ denote a \textit{backup} policy designed to bring the system inside a set $\XSetFI \subseteq \XSetSafe$ that is forward invariant under $\pibackup$. Then, MPS chooses $\ujoint_0 \in \{ \pijoint(\xjoint_0), \pibackup(\xjoint_0) \}$ and assigns $\ujoint_k = \pijoint(\xjoint_k)$ for $k > 0$ \cite{li2020robust,bastani2021safe}.
If taking $\ujoint_0 = \pijoint(\xjoint_0)$ does not lead to a feasible solution (i.e., $\xjoint_N \in \XSetFI$), then $\ujoint_0$ is taken to be $\pibackup(\xjoint_0)$.

MPS is extended to the multi-agent case in \cite{zhang2019mamps}, where safety under $\pibackup$ is considered agent-wise as opposed to for the entire MAS. This helps avoid suboptimal cases where all agents are forced to use $\pibackup$ even when only a single agent is not safe under $\pijoint$. However, \cite{zhang2019mamps} requires a centralized node to compute the MPS, and hence may have challenges scaling to a larger number of agents.

In \cite{vinod2022safe}, a PSF-like shield is used without the terminal state constraint \eqref{eq:psf:term_constr} for the \textit{joint} MAS. Linear dynamics and linear constraints are considered, which, along with no terminal constraints, allows for \eqref{eq:psf:optim} to be solved efficiently as a QP. Although removing the terminal state constraint also removes recursive feasibility guarantees, infeasibility was not reported in \cite{vinod2022safe} during hardware experiments.

Finally, in \cite{muntwiler2020distributedmpc}, a distributed method of PSF for linear systems with linear safety constraints and bounded disturbances is introduced. The disturbances are handled using a robust distributed MPC technique that uses tube MPC \cite{conte2013robust}, and a distributed negotiation procedure is introduced to allow agents to trade safety margins with neighbors. 

\subsection{Hamilton-Jacobi-based shielding}

The HJ methods are a class of optimal control tools for finding the reachable set of a dynamical system under worst-case disturbances; \cite{bansal2017hjreachability} provides a good introduction to this field. Most HJ methods proceed by solving a partial differential inequation for a scalar field over the joint state. This scalar field is known as the \textit{HJ value function} and its zero superlevel set is the control invariant set. The HJ value function also defines a controller that renders the system safe. A common HJ shielding strategy is to use this controller as a shield that only activates if the value function gets too close to zero~\cite{bansal2017hjreachability}. This shielding method can lead to undesirable bang-bang control behavior. There are other formulations that produce smoother HJ shielding controllers~\cite{choi2021cbvf}. The HJ-based methods have also been used to guide the learning of CBF controllers~\cite{tonkens2022cbfhj}.

A classic weakness of HJ methods is that they require solving a partial differential inequation over the state space of the system, and so the computational and memory requirements scale exponentially with the dimension of the state of the system~\cite{bansal2017hjreachability}. The memory requirements can be reduced by using function approximations, such as neural networks, to represent the HJ value function~\cite{fisac2019bridging,bansal2020deepreach}. However, this dependence on dimensionality nevertheless prevents HJ methods from being applied to the joint state of MAS. Instead, these methods factor the MAS into pairs of agents and then solve a pairwise HJ problem~\cite{chen2015hj}.

\subsection{Automata-based shielding}
Shielding has also been applied using tools from the field of \textit{formal methods}, where safety requirements are defined using linear temporal logic \cite{pnueli1977temporal}.
Shielding for safe learning-based control using automata was introduced in a single agent case in \cite{alshiekh2018safe}, borrowing ideas from \cite{bloem2015shield}, where a safety game \cite{mazala2002infinite} is solved to compute a set of states from where safety can be preserved.

In \cite{elsayed2021safe}, this is extended to the multi-agent case. Instead of constructing a single shield that monitors all agents, the state-space is decomposed into multiple pieces and a shield is constructed for each piece. The authors show that this allows the algorithm to scale from two to four agents on a grid world.
This work is extended in \cite{xiao2023model}, where this decomposition occurs dynamically. In \cite{melcer2022shield}, a  decentralized shield is constructed, improving scalability.

However, the shield synthesis tools used in these works require a finite abstraction of the state and control spaces \cite{bloem2015shield,alshiekh2018safe} and scale exponentially with the abstraction size \cite{konighofer2017shield}. This can lead to conservative behavior when coarse abstractions are used \cite{alshiekh2018safe}.

\subsection{Heuristic Projection}
Finally, there are shielding methods that do not provide formal safety guarantees but rather act as heuristics. In \cite{sheebaelhamd2021safe}, the heuristic approach from \cite{dalal2018safe} is used as a shield in a MARL framework. Here, the safety constraints are linearized, and it is assumed that only a single constraint is active at each time, giving rise to a closed-form solution that can be computed easily. In \cite{khan2019learning}, the velocity obstacle approach \cite{fiorini1998motion} is used as a shield. Here, agents velocities that are inside the velocity obstacle are projected back to the safe set. For a dynamic obstacle moving with constant velocity, a velocity obstacle defines the set of velocities of the agent that results in a collision with the dynamic obstacle \cite{fiorini1998motion}. When other agents are modeled as dynamic obstacles, the constant velocity assumption may not hold, but variants that make similar assumptions have been used successfully in practice \cite{snape2011hybrid}. 

The tradeoff between improved ease of use and computational cost is that these methods do not have the same safety guarantees as the previous categories of shields.

\section{Learning control barrier functions for MAS}\label{sec: cert learning}

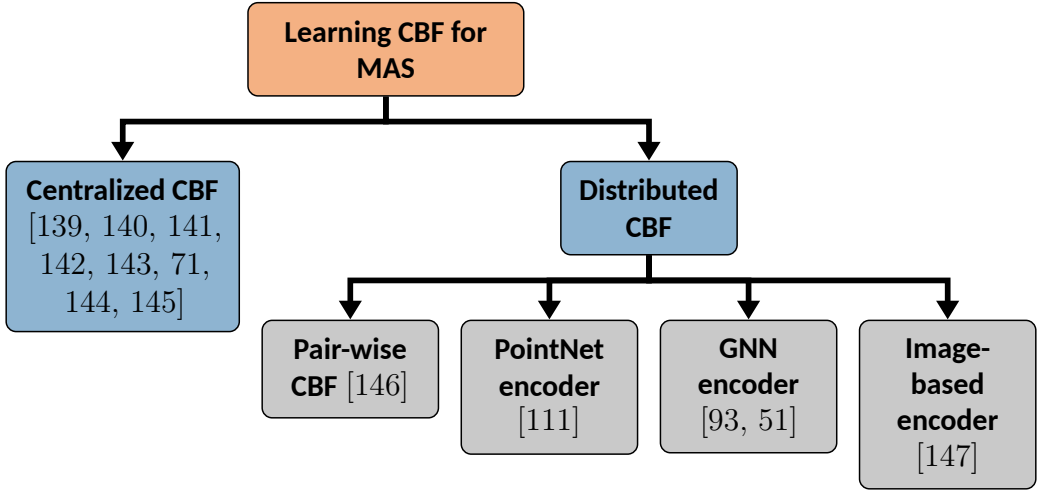
\begin{figure}
\centering
\begin{overpic}[width=\linewidth,abs,unit=1mm,scale=.25]{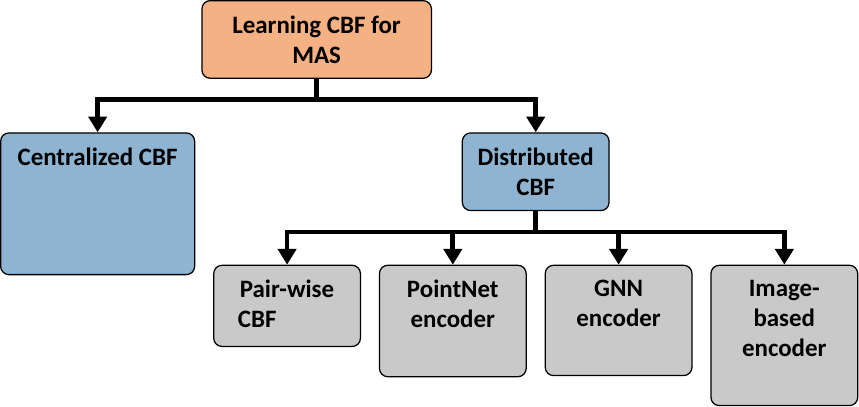}
  \put(2.9,33.8){\parbox[t]{26mm}{\centering\cite{saveriano2019learning,srinivasan2020synthesis,jin2020neural,robey2020learning,peruffo2021automated,dawson2022clbf,dai2022learning,yu2023sequential}}}
  \put(45.2, 12.6){\cite{meng2021reactive}}
  \put(68, 7.9){\cite{qin2020learning}}
  \put(92.3, 7.95){\cite{yu2023learning,zhang2023neural}}
  \put(121, 3.2){\cite{cui2022learning}}
\end{overpic}
\caption{Overview of learning CBF for MAS}
\label{fig:cbf_overview}
\end{figure}

Shielding is a powerful technique to guarantee the safety of the MAS. However, hand-crafting a shield is generally difficult, requires domain expertise, and can be done for a relatively small class of problems. Furthermore, it is computationally heavy to find a shield through optimization, especially for large-scale MAS with complex dynamics. To address this problem, there has been a lot of development on using machine learning to find a shield and use it as a guide for controller synthesis. Most of these works focus on a specific kind of shield, namely, CBFs \cite{dawson2022safe}, and tend to compute a CBF and a safe control policy simultaneously. In this manner, the computed control policy is encouraged to satisfy the CBF constraints and can be used for satisfying the safety requirements. In this section, we review approaches that learn a CBF and a control policy, for both centralized and distributed settings (see Table \ref{fig:cbf_overview} for an overview of learning CBF methods). 

\subsection{Centralized CBF}\label{sec: learn-cbf}

There is a plethora of work on learning centralized CBF for safety \cite{saveriano2019learning,srinivasan2020synthesis,jin2020neural,robey2020learning,peruffo2021automated,wang2023learning,xiao2023barriernet,wang2023physics,qin2022sablas}. The general idea of these approaches is to use \emph{self-supervised learning} for learning a common CBF for the entire MAS \cite{dawson2022safe}. To this end, first, a centralized CBF $B_\theta: \mathcal X\rightarrow\mathbb R$ and a centralized control policy $\boldsymbol{\pi}_\phi: \mathcal X\rightarrow\mathcal U$ are parameterized using Neural Networks (NNs) with parameter $\theta$ and $\phi$, respectively. Next, a loss function is designed to map the CBF constraint \eqref{eq: cbf-descent-cond} to a penalty term: 
\begin{align}
    \mathcal{L}_\mathrm{deriv}(\theta,\phi)&=\frac{1}{N_\mathrm{sample}}\sum_{\mathbf x\in\mathcal X}\left[-\frac{\partial B_\theta}{\partial \mathbf x}\mathbf F(\mathbf x,\boldsymbol{\pi}_\phi(\mathbf x))-\alpha\left(B_\theta(\mathbf x)\right)\right]^+,
\end{align}
where $[\cdot]^+=\max(\cdot,0)$ denotes the \texttt{ReLU} function, and $N_\mathrm{sample}$ is the total number of state samples collected. To render the safety of the system, following \Cref{rmk: cbf-safety}, it is essential that the $0$-superlevel set of $B_\theta$ is a subset of the safe set $\mathcal S$. However, ensuring such a condition is not straightforward. 
Therefore, one cannot na\"ively sample from the safe space and constrain the value of $B_\theta$ on these samples to be nonnegative. To design self-supervised learning losses, most works consider an alternative way: making $B_\theta$ negative everywhere in the unsafe space $\mathcal X\setminus\mathcal S$ \cite{srinivasan2020synthesis,jin2020neural}. Such a loss function can be readily defined as
\begin{equation}
    \mathcal{L}_\mathrm{unsafe}(\theta)=\frac{1}{N_\mathrm{unsafe}}\sum_{\mathbf x\in\mathcal{X}\setminus\mathcal S}\left[B_\theta(\mathbf x)\right]^+,
\end{equation}
where $N_\mathrm{unsafe}$ is the number of state samples in the unsafe space. However, if a loss of the form $\mathcal{L}_\mathrm{unsafe}(\theta)+\mathcal{L}_\mathrm{deriv}(\theta,\phi)$ is used, since there are only \textit{negative} samples, the learned policy can easily converge to a suboptimal solution where the forward-invariant set is relatively very small, if not empty. To address this problem, most of the work \cite{srinivasan2020synthesis,jin2020neural,dai2022learning,yu2023sequential} consider an additional loss term:
\begin{equation}
    \mathcal{L}_\mathrm{safe}(\theta)=\frac{1}{N_\mathcal{X_{\mathcal C}}}\sum_{\mathbf x\in\mathcal X_{\mathcal C}}\left[-B_\theta(\mathbf x)\right]^+,
\end{equation}
where $\mathcal X_{\mathcal C}\subset\mathcal S$ is the forward invariant set, and $N_\mathcal{X_{\mathcal C}}$ is the number of state samples in the forward invariant set, which act as \textit{positive} samples. Since the actual forward invariant set is unknown and computationally expensive to find, researchers generally approximate the set $\mathcal{X}_\mathcal{C}$ through various methods, such as using the set of initial conditions if it is known to be part of the forward-invariant set \cite{jin2020neural}, or using distance-based heuristic methods \cite{srinivasan2020synthesis,yu2023sequential}. 

Furthermore, to avoid learning a \textit{flat} CBF whose value is close to $0$ over the entire state space, many works also add a small margin $\nu>0$ in the loss terms \cite{dawson2022clbf}. Some works also consider a reference nominal controller $\boldsymbol{\pi}_\mathrm{nom}$ for the liveness property, such as reaching a goal location. As a result, another loss term $$\mathcal{L}_\mathrm{ctrl}=\frac{1}{N}\sum_{\mathbf x\in\mathcal X}\|\boldsymbol{\pi}_\phi(\mathbf x)-\boldsymbol{\pi}_\mathrm{nom}(\mathbf x)\|^2,$$ is added so that the learned controller is as close to the nominal controller as possible \cite{dawson2022clbf}. Note that the safety and the liveness properties encoded here via the means of different loss terms compete with each other in learning and often lead to a sub-optimal solution which either has a high performance with poor safety rate (learned controller close to the nominal controller) or high safety rate with poor performance \cite{zhang2023neural}.

Based on these individual loss terms, the parameterized CBF $B_\theta$ and the control policy $\boldsymbol{\pi}_\phi$ are trained using the loss function defined as
\begin{equation}
    \mathcal{L}_\mathrm{CBF}(\theta,\phi)=\mathcal{L}_\mathrm{deriv}(\theta,\phi)+\mathcal{L}_\mathrm{unsafe}(\theta)+\mathcal{L}_\mathrm{safe}(\theta) + \mathcal{L}_\mathrm{ctrl}(\phi).
\end{equation}
Upon convergence, a CBF and a safe control policy are obtained.  The training data for such learning methods is either collected by random sampling in the state space \cite{jin2020neural} or from simulated trajectories \cite{yu2023sequential,dai2022learning}. These approaches are more often that not generalizable to a large class of dynamics and relatively high-dimensional systems in contrast to the limited applicability of non-learning approaches. Also, these methods propose to learn a safe control policy along with the CBF. As a result, there is no need to solve the CBF-QP \eqref{eq: cbf-qp} during the execution, enabling them for real-time implementation. However, since the learned CBF is a neural network, it is difficult, if not impossible, to verify that the learned CBF candidate satisfies the CBF conditions \eqref{eq: cbf-descent-cond} \textit{everywhere} in the state space. In addition, the forward invariant set $\mathcal{X}_{\mathcal C}$ used in training is not easy to find. Moreover, as a centralized approach, it still needs global information and hence, it is not scalable for large-scale MAS \cite{qin2020learning,zhang2023neural}. 

\subsection{Distributed CBF}

In order to eliminate the need for global information and address the limited scalability of centralized CBF learning methods, researchers are now focusing on distributed CBF approaches\cite{qin2020learning,meng2021reactive,yu2023learning,zhang2023neural}. These methods assume that agents have access only to local observations and communication within their immediate vicinity.
Most of the works suppose that the agents are identical and share the same CBF and control policy. Thus, they use NNs to parameterize \textit{one} CBF $B_\theta$ and \textit{one} control policy $\pi_\phi$ that can be used for each agent \cite{qin2020learning,yu2023learning,zhang2023neural}, or each pair of agents \cite{meng2021reactive}, and use a similar procedure as discussed in \Cref{sec: learn-cbf} for learning them. Different from learning centralized CBF works, distributed CBF learning works often adopt the centralized training and distributed execution (CTDE) framework. During centralized training, the agents are trained jointly and the loss of agent $i$'s CBF can be backpropagated to its neighbors $j\in\mathcal N_i$ so that the distributed CBF conditions \eqref{eq: distributed-cbf-descent} are satisfied for all the agents \cite{zhang2023neural}. During distributed execution, the agents apply the learned controller which only uses the local observations to obtain control inputs. 

In contrast to the centralized CBF, where the central computation node has global observation, each agent only has local observation in distributed CBF. Different methods have been used to encode the local observation. For example, \cite{qin2020learning} uses a PointNet \cite{qi2017pointnet} so that the observation encoding is permutation-invariant to the observed agents. The recent work \cite{zhang2023neural} proposes to use graph neural networks (GNNs) with attention mechanism \cite{li2019graph} to encode the local observation. It utilizes the fact that GNNs with attention can handle a changing number of neighbors. The attention mechanism also addresses the problem of abrupt changes in the CBF value when an agent enters or leaves the sensing region of another agent. For image-based observations, convolutional neural networks (CNNs) and Long Short-Term Memory (LSTM) are used for encoding \cite{cui2022learning}.

Since learning a CBF for a large-scale MAS requires sampling from a large state space, it is not computationally tractable to explore the complete state space during learning and hence, the safety rate during execution is generally very low. To this end, an online policy refinement technique is applied during execution to obtain better safety performance \cite{qin2020learning,zhang2023neural}. The online policy refinement step adds a runtime gradient descent process to update the NN controller during execution. At any time step, if the learned control input does not satisfy the CBF descent condition, then the residue $\delta=[-\dot B - \alpha(B)]^+$ is computed, and gradient descent is used to update the learned control to minimize this residue. This is a distributed approach since computing $\dot B$ needs the knowledge of the control inputs of the neighboring agents also. 

Trained with a few tens of agents, the learned distributed CBF has demonstrated impressive generalizability in very large-scale systems constituting thousands of agents \cite{qin2020learning,zhang2023neural}. These approaches are also capable of using realistic and noisy LiDAR observation \cite{zhang2023neural} or image-based pixels \cite{cui2022learning}, instead of assuming that the \textit{actual} relative states are available. While these methods have several advantages as listed above, the learned CBF is hard to verify for correctness, and cannot provide formal guarantees on the safety of the MAS. Another challenge for these myopic distributed methods is deadlocks\cite{cohen2020approximate,reis2020control}, thereby compromising on liveness properties. 

\section{Safe multi-agent reinforcement learning}\label{sec: MARL}
Instead of delegating satisfaction of safety constraints to shielding-based methods or a learned CBF, one can also directly learn a policy to satisfy the safety constraints. One popular method of learning such a control policy is reinforcement learning (RL). In recent years, many of RL's biggest successes have been in its ability to play multi-agent games ranging from two-agent zero-sum games such as Go \cite{silver2016mastering,silver2017mastering} and Shogi \cite{schrittwieser2020mastering}, multi-agent zero-sum games such as Poker and Siplomacy, and team-based games such as Starcraft \cite{vinyals2019alphastar}, Dota \cite{berner2019dota}, Honor of Kings\cite{ye2020towards} and Football\cite{liu2021policy}. 

Despite these successes, there have been relatively few works that explicitly examine safety in Multi-Agent RL (MARL). The numerous approaches to both single-agent and multi-agent reinforcement learning address safety specifications by either terminating the episode when the safety specification is not met (e.g., in \cite{brockman2016openai}) or by including reward terms that discourage the \textit{unsafe} behavior (e.g., in \cite{brockman2016openai,tassa2018deepmind,hwangbo2019learning}).
While these approaches do not provide safety guarantees and can require reward function tuning, they are more popular in comparison to methods that explicitly address safety constraints.

\subsection{Flavors of safety in RL}
In the single-agent RL setting, a Constrained Markov Decision Process (CMDP) \cite{altman1999constrained} is the most popular problem formulation that additionally captures safety constraints. A MDP (Markov Decision Process) is defined as the tuple $(\mathcal{X}, \mathcal{U}, \mathbb{P}, r, \rho_0, \gamma)$, where $\mathbb{P}$ denotes the state transition probability, $r : \mathcal{X} \times \mathcal{U} \times \mathcal{X} \to \mathbb{R}$ denotes the reward function, $\rho_0$ denotes the starting state distribution, and $\gamma$ denotes the discount factor \cite{puterman2014markov}.
In the context of MAS, the reward function may encode liveness properties as in \ref{subsubsec:liveness}, such as goal reaching \cite{chen2017decentralized,zhang2019mamps,elsayed2021safe}
A CMDP $(\mathcal{X}, \mathcal{U}, \mathbb{P}, r, \rho_0, \gamma, C, d)$ extends a MDP by also considering a constraint function $C : \mathcal{X} \to \mathbb{R}$ and constraint bound $d \in \mathbb{R}$.
A CMDP seeks a policy that satisfies the \textit{average cost} constraints
\begin{equation} \label{eq:cmdp:average_cost}
    \sum_{k=0}^\infty \gamma^k C(x_k) \leq d.
\end{equation}
While average cost constraints \eqref{eq:cmdp:average_cost} are different from the safety constraints in Definition \ref{def:mas_safety}, the average cost constraints can be viewed as \textit{chance constraints} when $\gamma=1$, where $C$ is taken to be an indicator function $x \mapsto \mathbb{1}\{x \not\in \mathcal{S}\}$ and $d \in [0, 1]$ denotes the probability threshold, as
\begin{equation}
    \sum_{k=0}^\infty \mathbb{1}\{x_k \not \in \mathcal{S}\} = \Pr(x \not\in \mathcal{S}).
\end{equation}
On the other hand, for general choices of $C$ and $d$, this becomes a different notion of safety than Definition \ref{def:mas_safety}, and the optimal solution of the CMDP could result in exiting the safe set $\mathcal{S}$.
Single-agent RL methods that explicitly solve the CMDP label themselves as constrained RL or safe RL (e.g., \cite{altman1999constrained,gu2022review,yu2022reachability}) and are commonly tested on benchmarks such as safety gym \cite{ray2019benchmarking}.

Another notion of safety is when no constraint violation is allowed at any \textit{any} timestep $k$, i.e.,
\begin{equation} \label{eq:cmdp:peak}
    \max_{k \geq 0} C(x_k) \leq d
\end{equation}
This is known as a peak constraint \cite{geibel2005risk,geibel2006reinforcement} or state-wise safety \cite{zhao2023state} in the Safe RL literature, and aligns with the notion of safety considered in this paper, where the safe set $\XSetSafe$ is defined as
\begin{equation}
    \XSetSafe \coloneqq \{ x \in \XSet | C(x) \leq d \}.
\end{equation}
While peak constraints are not as common for constrained RL methods, there are some works that tackle this problem \cite{fisac2019bridging,yu2022reachability,so2023solving} (see \cite{zhao2023state} for a recent survey of some methods).
It can be related to the average cost case \eqref{eq:cmdp:average_cost} by taking $\tilde{C_k} \coloneqq \max(0, C_k - d)$ and $\tilde{d} = 0$ \cite{so2023solving}, yielding the constraints
\begin{equation}
    \sum_{k=0}^\infty \gamma^k \max(0, C_k - d) \leq 0
\end{equation}
For both types of constraints, taking smaller values of $d$ results in a stricter constraint on the total accrued cost and hence a lower total reward, while higher values of $d$ loosen the constraint and allow higher rewards.

\subsection{Safe MARL}
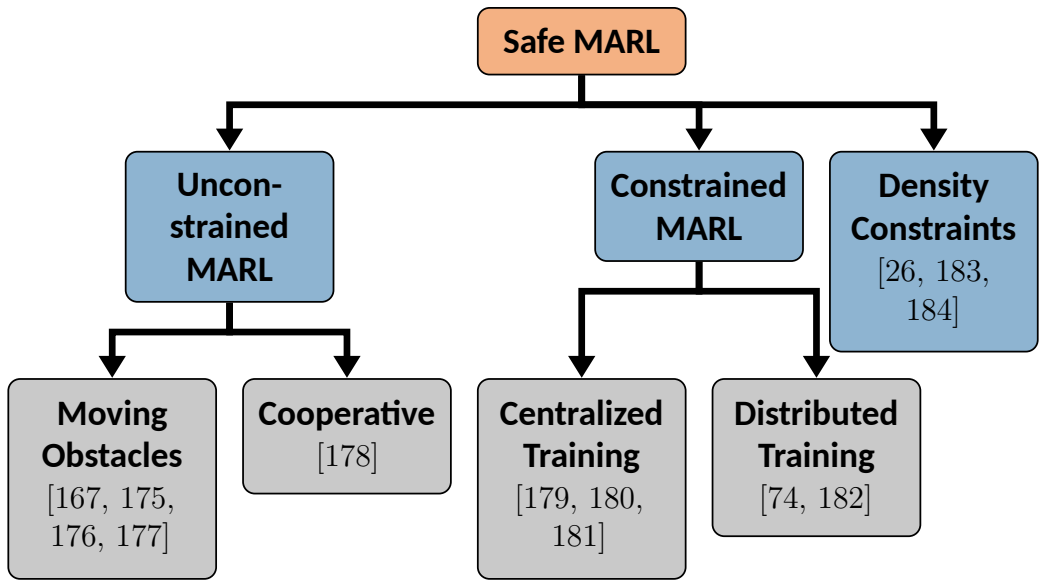
\begin{figure}
\centering
\begin{overpic}[width=\linewidth,abs,unit=1mm,scale=.25]{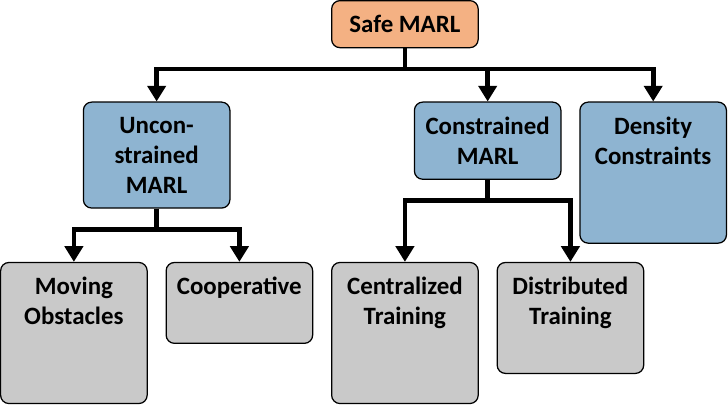}
  \put(4.7, 9.6){\parbox[t]{18mm}{\centering\cite{chen2017decentralized,chen2017socially,everett2018motion,semnani2020multi}}}
  \put(41, 15.2){\cite{lowe2017multi}}
  \put(67, 9.6){\parbox[t]{18mm}{\centering\cite{liu2021cmix,ding2023provably,gu2023safe}}}
  \put(99.9, 9.6){\cite{lu2021decentralized,geng2023reinforcement}}
  \put(114.1, 39.7){\parbox[t]{18mm}{\centering\cite{liu2021policy,chen2023density,qin2021density}}}
\end{overpic}
\caption{Overview of Safe MARL-based approaches}
\label{fig:marl_overview2}
\end{figure}

Given that solving the CMDP problem and obtaining policies that successfully satisfy the safety constraints is already challenging in the single-agent case, it is even more challenging in the MAS setting. Consequently, there have been relatively fewer works that tackle the problem of safe MARL. This is noted in many surveys on MARL, which mention safe MARL to be a direction that is relatively unexplored \cite{oroojlooy2023review,zhang2021multi,gu2022review,yang2020overview}. Nevertheless, recent years have seen an increasing number of works that tackle this challenging problem. We provide an overview of existing methods in
Figure \ref{fig:marl_overview2}
, which we describe in more detail in the coming subsections.

\subsection{Unconstrained MARL}
Early works that approached the problem of safety for MARL focused on navigation problems and collision avoidance \cite{chen2017decentralized,chen2017socially,everett2018motion,semnani2020multi}, where safety is achieved by either a sparse collision penalty \cite{long2018towards}, or a shaped reward term that penalizes small distances to obstacles and neighboring agents \cite{chen2017decentralized,chen2017socially,everett2018motion,semnani2020multi}.
However, satisfaction of collision avoidance constraints is not necessarily guaranteed by either the final policy or even the optimal policy \cite{massiani2022safe}. Consequently, while these methods report $100\%$ safety rates empirically when there are fewer agents, safety violations occur when tested with larger number of agents (e.g., $>3\%$ for $8$-agents in \cite{everett2018motion}, $>3\%$ for $20$-agents in \cite{long2018towards}).
These trends are also similar for liveness properties. In \cite{everett2018motion}, although the use of reward to encourage goal-reaching is suffucient for all agents to eventually reach their goal when there are less than $3$ agents, the percentage of deadlocks increases to $1.6\%$ for $10$ agents. Improving satisfaction of the liveness properties may require alternate approaches such as imitation learning of Multi-Agent Path Finding (MAPF) algorithms \cite{sartoretti2019primal,damani2021primal}, where success rates are close to $100\%$ even for up to $128$ agents. 

While, there exists some finite scale for the penalty such that the optimal policy is guaranteed to satisfy the constraints \cite{massiani2022safe},
too large of a penalty term often results in poorer performance empirically \cite{shalev2016safe}.  
As noted in \cite{shalev2016safe}, this can be explained by larger penalties resulting in larger variances in the reward which ultimately results in poorer optimization performance.

\subsection{Constrained MARL}
In contrast to unconstrained MARL methods that penalize safety violations in the reward term and then solve the resulting unconstrained problem, constrained MARL methods explicitly solve the CMDP problem.
For the single-agent case, prominent methods for solving CMDPs include primal-dual methods using Lagrange multipliers \cite{borkar2005actor,tessler2018reward} and via trust-region-based approaches \cite{achiam2017constrained}. These methods provide guarantees either in the form of asymptotic convergence guarantees to the optimal (safe) solution \cite{borkar2005actor,tessler2018reward} using stochastic approximation theory \cite{robbins1951stochastic,borkar2009stochastic}, or recursive feasibility of intermediate policies \cite{achiam2017constrained,satija2020constrained} using ideas from trust region optimization \cite{schulman2015trust}.
The survey \cite{gu2022review} provides an in-depth overview of the different methods of solving safety-constrained single-agent RL.

MARL algorithms can be broadly divided into two paradigms: centralized and distributed training algorithms \cite{gronauer2022multi}. The same holds true for their extensions to consider safety constraints.  

\subsubsection{Centralized training}
During centralized training, agent policies are updated at a central node, where information in addition to each agent's local observations is used to update each agent's policies. This is the dominant paradigm for unconstrained MARL \cite{gronauer2022multi} and is referred to as Centralized Training Decentralized Execution (CTDE). 

One of the first safe CTDE methods to be proposed is in \cite{gu2023safe}, where the authors combine Constrained Policy Optimization (CPO) \cite{achiam2017constrained}, a method for solving CMDPs, with Heterogeneous-Agent Trust Region Policy Optimisation (HATRPO), a
MARL method that enjoys a theoretically-justified monotonic improvement guarantee \cite{kuba2022trust}. Theoretical analysis guarantees monotonic improvement in reward while
theoretically satisfying safety constraints during each iteration, assuming that the initial policy is feasible, the value functions are known and a trust-region optimization problem can be solved exactly. However, neither of these assumptions are guaranteed in the method implemented in practice due to approximation errors in the value function and a quadratic approximation of the trust-region problem \cite{gu2023safe,kuba2022trust}.

Another early safe CTDE method is CMIX \cite{liu2021cmix}, which extends the value function factorization method QMIX \cite{rashid2020monotonic} to additionally consider both average constraints and peak constraints.
However, no theoretical analysis on the convergence of the proposed algorithm is given.
\cite{liu2021cmix,ding2023provably,gu2023safe}

\subsubsection{Distributed training}
In distributed training methods, each agent has a private reward function, policy updates occur locally for each agent, and communication between agents is used to arrive at a policy that minimizes the total reward subject to safety constraints \cite{lu2021decentralized}.
In this distributed setting where the reward and constraints are private and each agent has a different policy when the algorithm has not converged, it is not possible to evaluate the performance of each agent's policy. Consequently, these approaches are based on primal-dual optimization and provide asymptotic convergence guarantees to local optima \cite{lu2021decentralized}, but are unable to provide any safety guarantees before convergence.

\subsection{Density-based approaches}
Finally, a vastly different approach to safe MARL looks at the case when the agents are indistinguishable, the number of agents is very large  and applies the mean field approximation, where they look at the limit as the number of agents goes to infinity, and the quantity of interest is instead the density of the overall swarm of agents \cite{lasry2007mean,bensoussan2013mean}.
As the concept of individual agents is gone, both interagent and agent-obstacle constraints are replaced by density constraints \cite{lin2021alternating,liu2022deep}.
Navigation problems when applying the mean field approximation can be viewed as an optimal transport problem \cite{liu2022deep}. When also considering safety constraints via state-dependent costs, the problem becomes a mean field game with distributional boundary constraints that can be solved using RL techniques \cite{liu2022deep}.

For example, \cite{lin2021alternating,liu2022deep} consider a navigation problem with obstacles, where the density of the multi-agent swarm at the obstacles is constrained to be zero for a given initial and final distribution of agents. In \cite{liu2022deep}, a penalty on the density of agents is also included to incentivize agents to spread out more, and consequently reduce the risk of inter-agent collisions.

\section{Safety verification for learning-enabled MAS}\label{sec: MAS verification}

\begin{figure}[h]
  \centering
  \includegraphics[width=1\textwidth]{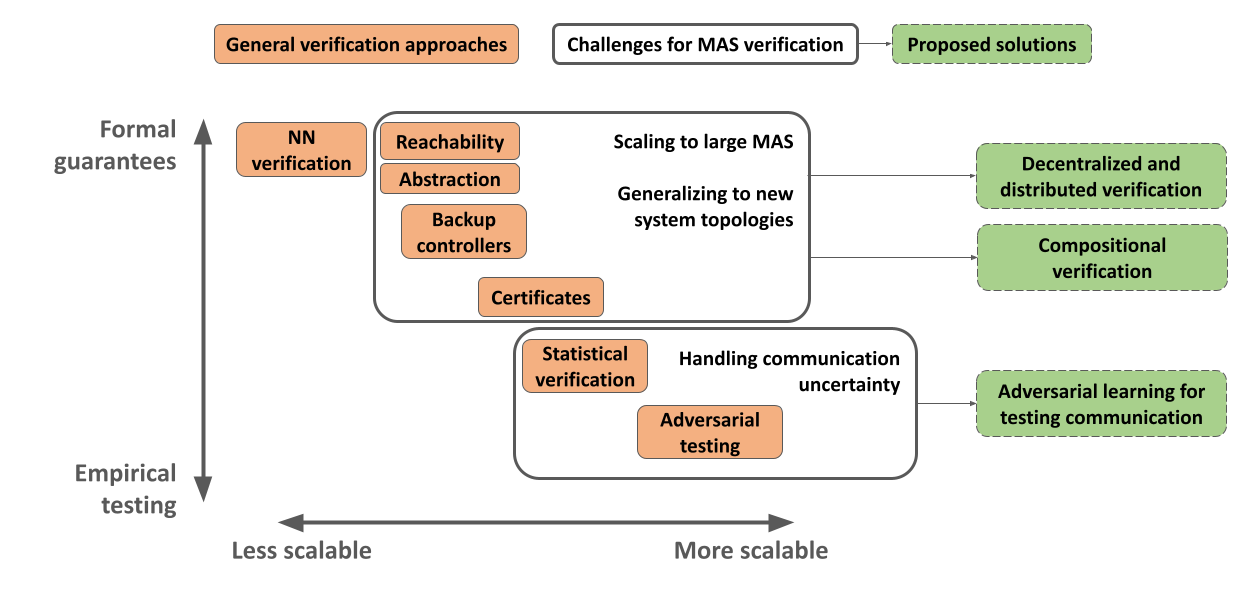}
  \caption{An overview of verification methods, comparing the ability to provide formal guarantees with the ability to scale in practice.}
  \label{fig:verification_tradeoff}
\end{figure}

Once a control policy has been learned, it must be checked for correctness before it can be deployed, particularly in safety-critical control contexts. This is particularly true for control policies represented using difficult-to-interpret models such as NNs. This section reviews methods for checking the safety properties of a learned controller or shield for MAS. Broadly speaking, these methods can be organized as shown in Fig.~\ref{fig:verification_tradeoff}, where the primary trade-off is between the ability to provide formal guarantees and the ability to scale to practically-sized problems. Fig.~\ref{fig:verification_tradeoff} also highlights the specific challenges of MAS verification and summarizes existing approaches to resolving these challenges.

We begin by reviewing the methods from Fig.~\ref{fig:verification_tradeoff} with reference to the extensive literature for single-agent verification, before highlighting the specific challenges that arise in the multi-agent setting. We then review methods for addressing each of these challenges, concluding with a discussion of open problems.

\subsection{Review of single-agent verification tools}

There are a number of excellent surveys for centralized verification that provide a good starting point for readers interested in a broad introduction to the field; we will review these surveys here, then devote most of our attention in this survey to the challenges that distinguish multi-agent verification from the centralized case.

The survey \cite{corso2022verificationsurvey} covers search- and optimization-based methods for black-box autonomous systems. These methods can sometimes provide formal guarantees, depending on the completeness of the underlying optimizer or search algorithm; for example, some stochastic global optimization methods enjoy asymptotic completeness guarantees under certain assumptions, which may be used to provide formal guarantees, but most black-box optimization methods are best suited for empirical testing. In this survey, we discuss the challenges in scaling these methods to large-scale MAS, as well as how search- and optimization-based methods can be extended to consider issues that are specific to MAS, such as communication noise and cybersecurity.

Centralized reachability analysis using Hamilton-Jacobi (HJ) methods are covered in the survey \cite{bansal2017hjreachability}, which are specialized for checking set invariance properties but can provide formal guarantees. The survey \cite{bansal2017hjreachability} discusses centralized verification of MAS (i.e. collapsing the MAS into a single dynamical system) and verification of pairwise safety (i.e. checking for inter-agent collision avoidance), but they do not consider MAS safety more generally. Other methods of centralized reachability analysis (e.g., set propagation), have been extensively covered in previous surveys \cite{alur2011formal,althoff2021set,chen2022reachability}. These methods use sets to overapproximate the reachable set of a dynamic system and can be used to provide safety guarantees. However, these surveys do not focus on the topics of learned controllers or the reachability of MAS. We restrict our discussion to decomposition-based approaches to scaling HJ methods, along with other reachability analysis tools, for MAS.

Another recent review article \cite{dawson2022safe} provides a survey of neural certificates, but they focus largely on the centralized case. We extend the discussion of certificates to consider decentralized, distributed, and compositional approaches checking to certificates. Due to the extensive coverage of certificate learning in Section \ref{sec: cert learning}, in this section we only discuss the challenges involved in checking the validity of these certificates in the multi-agent case.

In \cite{liu2021nnverification}, the authors provide a survey of techniques for verifying input-output properties of neural networks, some of which have been applied to verify the soundness of shields like barrier and Lyapunov functions~\cite{abate2021fossil,dai2021lyapunovstable}. Due to the extreme fundamental complexity of neural network verification (an NP-complete problem~\cite{liu2021nnverification}) and several open technical challenges that we discuss in the sequel, these methods have yet to be applied to MAS.

Similarly, other surveys cover verification for safe learning and control~\cite{brunke2022safe} and autonomous aerial systems~\cite{brat2023autonomy}; while these provide a good overview of their respective fields, they do not specifically address MAS.

\subsection{Challenges of MAS verification}\label{sec: challenges MAS verification}

While many of the concepts from centralized verification extend to the multi-agent case, there are a number of challenges that are unique to MAS. We discuss these challenges here, and the rest of this section is devoted to reviewing proposed methods for addressing these issues.

\subsubsection{Scalability to high dimensions}

The most immediate challenge in verifying learned control systems for MAS is the scale. Formal verification using methods like HJ~\cite{bansal2017hjreachability} and abstraction~\cite{alur2000abstraction} are notoriously difficult to scale to systems with large state dimension, so verifying an MAS by na\"ively concatenating the states of individual agents to form a centralized verification problme is generally not tractable. Similarly, neural network verification is generally intractable for networks with more than a few hundred neurons~\cite{liu2021nnverification}, so the formal guarantees that might be derived using these methods are hard to apply in practice without additional decomposition of the problem.

Even for empirical testing methods, such as those relying on stochastic optimization, it can be challenging to efficiently explore high-dimensional space. The challenge of scalability has lead to a range of specialized techniques for MAS and other large-scale systems, including decentralized certificates, distributed search and optimization, and compositional verification methods, which we discuss in the following section.

\subsubsection{Handling changing system topology}

The second specific challenge for verifying MAS is that, unlike single-agent systems, MAS can be dynamically reconfigured at runtime. For example, the connectivity of agents might change during execution, or the exact number of agents in the system might be  uncertain at test-time. As a result, in order to be most useful for MAS, verification methods must be robust to both number and topology of agents in the system; in particular, formal guarantees should generalize to different system topologies, and empirical testing should achieve sufficient coverage of the range of topologies that we expect to see at runtime. In the following, we survey several methods that meet this requirement, such as compositional approaches and adversarial testing, and we identify several important open questions in this area.

\subsubsection{Handling communication delay and error}

An additional unique feature of MAS verification is that there is a class of disturbances --- those affecting inter-agent communication --- that are typically not present in single-agent settings, and thus typically not considered by single-agent verification methods~\cite{brunke2022safe,corso2022verificationsurvey}. Communication disturbances come in many forms, including both naturally occuring factors like communication delay and noisy or lossy communication and adversarial effects like non-cooperative or malicious agents that can send arbitrary messages to other agents (this latter category is important in considering the cybersecurity of MAS).

\subsection{Decentralized and distributed verification}

To address the first two challenges (scalability and generalizing to changing system configurations), a natural question is how centralized verification algorithms can be adapted to the distributed or decentralized setting. Decentralized~\cite{qin2020learning,wang2017safety} and distributed~\cite{zhang2023neural} barrier certificates are one example of this approach, which we discuss extensively in Sections~\ref{sec: shielding} and~\ref{sec: cert learning}. Other examples include decentralized reachability analysis, e.g. only considering pairwise interactions~\cite{bansal2017hjreachability,wang2020reachability},
and distributed optimization to guarantee safety via a multi-agent safe model-predictive control \cite{muntwiler2020distributedmpc}.

There are several important considerations for decentralized verification. First, it is not possible to verify some MAS safety and liveness properties in a fully decentralized setting. For example, obstacle avoidance (involving only individual agents and the environment) can be verified in a decentralized manner, but inter-agent collision avoidance requires considering pairs or small cliques of agents~\cite{bansal2017hjreachability,zhang2023neural,qin2020learning}. Other properties are more nuanced; for example, connectivity maintenance can be verified using only pairwise communication if the communication graph topology is fixed (this results in a pairwise maximum distance constraint), but if the topology is allowed to vary then agents must be able to compute eigenvalues of the communication graph Laplacian in a distributed manner~\cite{Cavorsi-RSS-22}.

A second consideration concerns the common strategy of guaranteeing safety by constructing a \textit{safety filter} using either barrier certificates~\cite{qin2020learning,wang2017safety,zhang2023neural} or reachability analysis~\cite{muntwiler2020distributedmpc}, as discussed in Section~\ref{sec: shielding}. This safety filter limits the range of actions that individual agents may take and often provides formal safety guarantees, but this limit often reduces the optimality of the filtered policy, and this reduced optimality can be more severe for decentralized safety filters. For example, \cite{chen2016masmiphj} shows that a centralized safe controller achieves a higher task completion rate than a controller that only considers pairwise interactions.

Finally, the application of existing neural network verification (NNV) tools to decentralized learned control policies or certificates remains an open problem; even if more performant NNV tools are developed that can scale to large policy networks, existing NNV algorithms cannot handle architectures like graph neural networks (GNNs) commonly used in learning for decentralized and distributed control~\cite{salzer2023fundamental}.

\subsection{Compositional verification methods}

An alternative to the distributed and decentralized methods discussed above is a more scalable form of centralized verification where agent-level guarantees are hierarchically composed to yield guarantees for the entire MAS; this class of approach is known as \textit{compositional verification}. Compositional certificates have been used to prove MAS stability properties by composing individual-agent stability certificates that are either learned~\cite{zhang23compositional} or found using sum-of-squares optimization~\cite{shen2018compositional}. The benefit of compositional approaches to certificate learning is that they require much less data to train; they can be trained at the individual agent level and then generalized to large number of agents to provide system-level guarantees~\cite{zhang23compositional}.

A similar compositional approach has been applied to reachability-based verification, for example using reachability to certify pairwise collision avoidance~\cite{bansal2017hjreachability}, which composes to give system-level inter-agent collision avoidance guarantees under certain assumptions about the sparsity of agents, or composing reachable sets for subsystems to certify properties of networked systems~\cite{althoff2014composition}. It is important to note that decomposition of system-level properties to the agent level necessarily constrains the space of safe control policies, potentially introducing conservatism; e.g. \cite{chen2016masmiphj} show that composing pairwise collision-free policies leads to more conservative behavior than a more general $N$-agent decomposition. An important open question in compositional verification is understanding how the choice of decomposition affects the conservatism (or even feasibility) of the resulting control strategy.

\subsection{Handling communication uncertainty}

A unique feature of MAS (compared to centralized systems) is that they often rely on communication links between agents to function. In a fully decentralized setting, agents might only require local observations of neighbors (e.g. relative position) without explicit communication, but distributed systems can rely on multiple rounds of message passing to achieve consensus or accomplish a distributed optimization task~\cite{molzahn2017survey,espina2020distributed}. If this communication is disrupted, or if adversarial agents inject malicious communication packets, then the safety of the overall MAS can be compromised.

A number of surveys consider the problem of communication uncertainty and cybersecurity for MAS; common strategies for ensuring robustness to communication issues include fault detection and isolation (FDI), secure consensus algorithms (e.g. where a non-majority subset of agents can be compromised while still maintaining the integrity of the non-compromised agents~\cite{zhang2021cyber}), and safety filters that maintain sufficient connectivity of the communication graph~\cite{Cavorsi-RSS-22}. Here, we review how these security methods can be extended using learning-based methods. For example, \cite{garg2023fdi} learns a model-free FDI policy for single-agent systems with multiple faults; this method can be extended for model-free learning-based FDI in MAS. Other works in this vein include using representation learning to classify the network robustness of MAS~\cite{wang2018mlnetwork} or using RL to detect intrusions in a networked system~\cite{servin2008marlid} or generate adversarial attacks on MAS~\cite{yamagata2021falsification}.

A related communication issue arising in MAS is delay, which in the worst case can prevent the system from achieving consensus or even destabilize the system~\cite{papachristodoulou2010delay}. Several methods have been proposed to handle communication effects, including delay and uncertainty, in the reinforcement learning literature~\cite{jiang2018attentionalcomm, wang2020informationbottleneck, das2019tarmac, kim2021communication}. However, verifying (either formally or empirically) the robustness of a learned controller to these effects remains a challenging open problem.

\section{Open Problems}\label{sec: open probs}
Given the state of the art reviewed in this survey, a few themes stand out as areas for future work. The rest of this section discusses these themes.

\subsection{Combined safety and liveness guarantees}

While learning-based methods for MAS have seen immense progress in the past decade, much work is still needed when it comes to safety, provable guarantees, and scalability. 
In the context of learning, there exists a trade-off between liveness properties, such as goal reaching, and safety properties, such as collision avoidance. It is still an open problem as to how to achieve high safety rates or provable safety guarantees along with high performance or guarantees on deadlock resolutions, especially in partially observable systems and while applying distributed algorithms \cite{qin2020learning,zhang2023neural}. 

\subsection{Decomposition}

For any single-agent method, one of the main challenges for their extension to MAS is how to perform a suitable decomposition that balances the performance and scalability of the method.

\paragraph{Appropriate choice of decomposition for shielding}
For shielding-based approaches, this decomposition has been done via distributed computation techniques to efficiently compute a centralized shielding function \cite{pereira2022decentralized}, factoring the state space \cite{elsayed2021safe,xiao2023model}, distributed decomposition of the safety constraints (i.e., \cite{zhang2019mamps}), or performing a completely decentralized factorization of the shielding function such that the communication is not needed \cite{cai2021safe,melcer2022shield}. A major open question with these approaches is to explore how the type of decomposition affects the conservatism of the resulting safety filter and its ability to scale and how different choices of shield type (e.g., PSF, CBF,\dots) can change this tradeoff between ease of construction, scalability, and conservatism.

\paragraph{Appropriate choice of decomposition for verification}
Similarly to shielding methods, verification methods also often rely on decomposition. An important open problem discussed in Section~\ref{sec: MAS verification} is how this choice of decomposition affects the conservatism and completeness of the resulting verification scheme. Furthermore, care must be taken to ensure that the decomposition is valid; for example, if a system is only analyzed pairwise for collision avoidance, how does the verification method account for conflicts between more than two agents?

\subsection{Practical issues}

\paragraph{Safety under communication uncertainty}
Another challenge in designing safe methods for MAS is handling dynamically changing MAS configuration, communication delays, package losses, and adversarial communication. Designing control synthesis and verification tools that are both scalable and robust to time-varying topology as well as communication-related issues remains an open problem, as many existing methods either ignore communication effects or rely on domain-specific information.

\paragraph{Strict requirements of CBF-QP}
While using CBF for shielding, a crucial practical issue is the feasibility of the CBF-QP. Real-world systems often have input constraints, e.g., torque limit, acceleration limit, etc.
The CBF-QP may become infeasible when input constraints are included.
Guaranteeing the feasibility of the CBF-QP is an open problem \cite{chen2020guaranteed,agrawal2021safe,cortez2021robust}, although it is likely that future work will address these issues.

\paragraph{Safe methods are complex and unpopular in practice}
When it comes to RL, one of the main challenges for safety in RL in both the single-agent and multi-agent cases is the tradeoff between the simplicity of the algorithm, practical performance, and safety guarantees. With unconstrained RL, in general, neither the resulting policy after training nor the theoretically optimal policy is guaranteed to satisfy the safety constraints \cite{massiani2022safe,tasse2023rosarl}.
On the other hand, while some constrained RL methods have convergence and safety guarantees (e.g., \cite{gu2021multi}), these methods have more components and thus are more difficult to implement and use in practice as compared to unconstrained variants. As a result, these safe methods are relatively unpopular among practitioners compared to unconstrained methods.

\paragraph{Value of methods without practical safety guarantees}
Another challenge for safety in MAS is the question of whether learned CBFs or safe MARL methods should be used instead of shielding-based methods when safety guarantees are desired. Although constrained RL can provide per-iteration safety guarantees, this assumes both an initially feasible policy and access to the true value function, neither of which is guaranteed to hold in practice \cite{gu2023safe}. Similarly, a learned CBF, if verified to be an actual CBF, inherits the theoretical safety guarantees. However, the training process for learning CBFs does not guarantee that this will always be true. On the other hand, shielding-based methods shift this complexity from the algorithm to the user, who must first construct a valid shielding function in order to provide provable safety guarantees during both training and deployment. 
It is not clear whether this tradeoff between scalability and practically relevant safety guarantees will be attractive for safety-critical applications.

\section{Conclusions}\label{sec: conclusions}
MAS are ubiquitous in today's world, with potential for applications ranging from robotics to power systems, and there is a large body of literature on control of MAS. However, the safe control design of large-scale robotic MAS is a challenging problem. In this survey, we have reviewed how various learning-based methods have shown promising results in addressing some of the aspects of safe control of MAS, such as the 
safety guarantees of shielding-based methods, the generalizability of
learning CBF
methods, and the wide applicability of MARL-based methods.
Despite their advantages, no existing method has all the desired properties of being provably safe, scalable, computationally tractable, and implementable to a variety of MAS problems. 
{While certificate learning and safe MARL-based methods can provide theoretical safety guarantees, these guarantees do not hold in practice due to unrealizable assumptions.}
We have identified a range of open problems covering these concerns, and we hope that our review of the state-of-the-art in this field provides a springboard for further research to address these issues and realize the full potential of safe learning-based control for MAS.

\newpage
\printbibliography

\end{document}